\documentclass{article}

    \usepackage[nonatbib,preprint]{neurips_data_2024}

\usepackage[utf8]{inputenc} 
\usepackage[T1]{fontenc}    
\usepackage{url}            
\usepackage{booktabs}       
\usepackage{amsfonts}       
\usepackage{nicefrac}       
\usepackage{microtype}      
\usepackage{xcolor}         
\usepackage{amssymb} 
\usepackage{colortbl}
\usepackage{color}
\usepackage{amsfonts}
\usepackage{amsopn}
\usepackage{enumitem}
\usepackage{makecell}
\usepackage{diagbox}
\usepackage{stfloats}
\usepackage{xspace}
\usepackage{multirow}
\usepackage{ulem}
\usepackage{mathrsfs}
\usepackage{algorithm}
\usepackage{algorithmicx}
\usepackage{algpseudocode}
\usepackage{array}
\usepackage{caption}
\usepackage{lipsum}
\usepackage{amsmath}
\usepackage{graphicx}
\usepackage{bbm}

\usepackage{wrapfig}

\usepackage{titlesec}
\titlespacing*{\paragraph}{0pt}{-0.5ex}{0.5em}

\definecolor{lightblue}{RGB}{0, 110, 204}
\usepackage[pagebackref=true,breaklinks=true,letterpaper=true,citecolor=lightblue,colorlinks,bookmarks=false]{hyperref}
\def\DatasetName{{Holistic-Motion2D}}
\def\MethodName{{Tender}}
\title{\raisebox{-0.2\height}{\includegraphics[height=1.0\baselineskip]{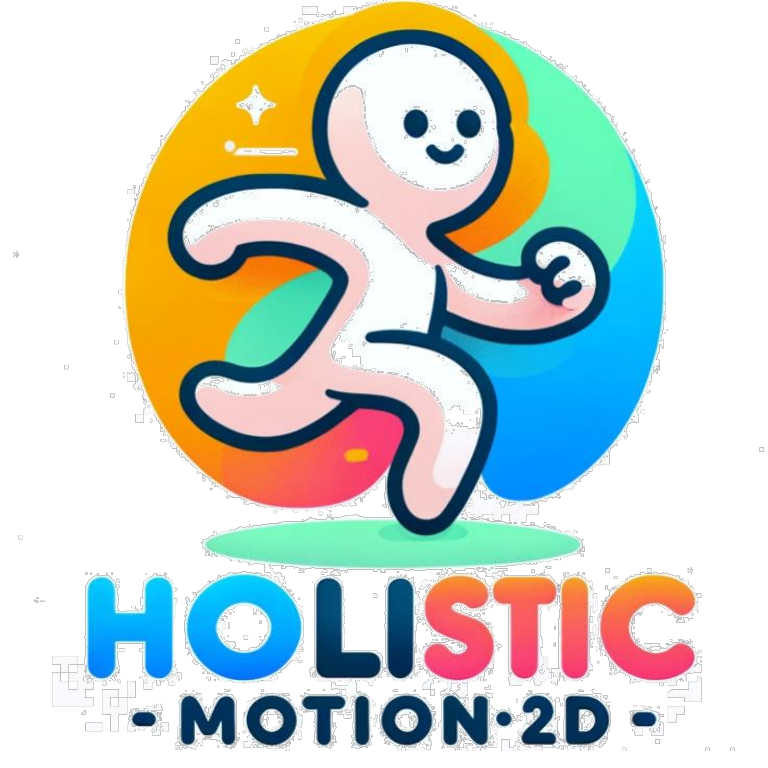}}Holistic-Motion2D: Scalable Whole-body \\ Human Motion Generation in 2D Space}

\author{%
\textbf{Yuan Wang}$^{1,5,6\ddagger}$ 
~~
\textbf{Zhao Wang}$^{2\ddagger}$ 
~~
\textbf{Junhao Gong}$^{3\ddagger}$ 
~~
\textbf{Di Huang}$^{4,5}$ 
~~
\textbf{Tong He}$^{5}$  \\
~~
\textbf{Oyang Wanli}$^{5}$ 
~~
\textbf{Jile Jiao}$^{1,6}$ 
~~
\textbf{Xuetao Feng}$^{1,6}$ 
~~
\textbf{Qi Dou}$^{2}$ 
~~
\textbf{Shixiang Tang}$^{2\clubsuit}$ 
~~
\textbf{Dan Xu}$^{7}$ 
~~
\\
$^{1}$Tsinghua University
~~
$^{2}$The Chinese University of Hong Kong 
~~
$^{3}$Shandong University \\
~~
$^{4}$The University of Sydney 
~~
$^{5}$Shanghai AI Laboratory 
~~
$^{6}$Alibaba Group
~~
$^{7}$HKUST \\
$^{\ddagger}$Equal Contribution 
~~
$^{\clubsuit}$Corresponding Author
~~
}

\begin{document}

\maketitle

\begin{figure}[h]
    \begin{center}
        \centerline{\includegraphics[width=\linewidth]{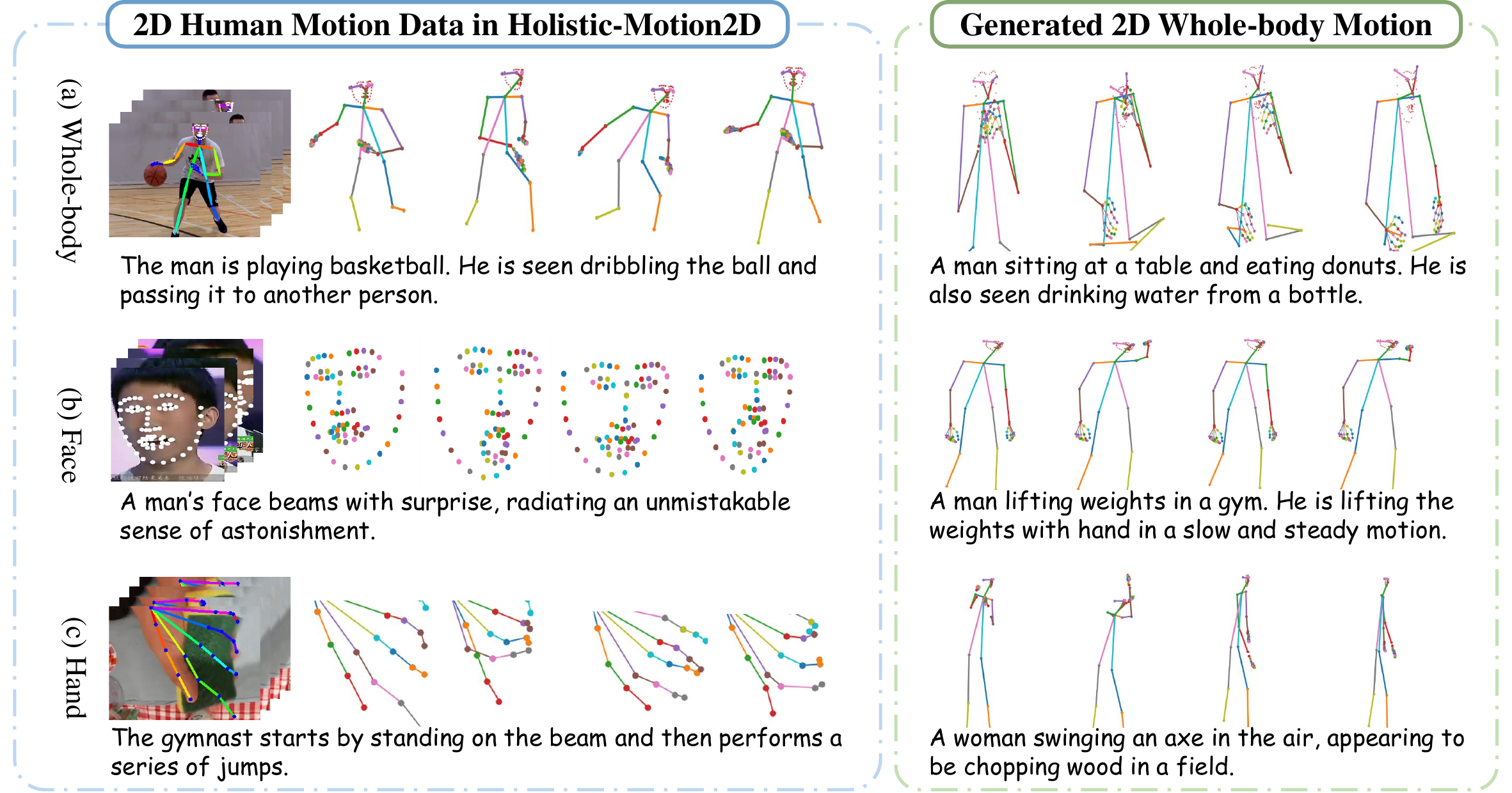}}
        \vspace{-2mm}
        \caption{{\bf Overview of \textbf{\DatasetName} and generated 2D whole-body motions.} {\bf Left:} 2D human motion data in our dataset with (a) whole-body, (b) face, and (c) hand motions. {\bf Right:} the generated 2D whole-body motion from our model. Every 2D motion sequence is shown following a temporal progression from left to right.} 
        \label{fig:teaser}
    \end{center}
\end{figure}

\begin{abstract}
In this paper, we introduce a novel path to \textit{general} human motion generation by focusing on 2D space.
Traditional methods have primarily generated human motions in 3D, which, while detailed and realistic, are often limited by the scope of available 3D motion data in terms of both the size and the diversity.
To address these limitations, we exploit extensive availability of 2D motion data.
We present \textbf{\DatasetName}, the first comprehensive and large-scale benchmark for 2D whole-body motion generation, which includes over 1M in-the-wild motion sequences, each paired with high-quality whole-body/partial pose annotations and textual descriptions.
Notably, Holistic-Motion2D is ten times larger than the previously largest 3D motion dataset. 
We also introduce a baseline method, featuring innovative \textit{whole-body part-aware attention} and \textit{confidence-aware modeling} techniques, tailored for 2D \uline{T}ext-driv\uline{EN} whole-bo\uline{D}y motion gen\uline{ER}ation, namely \textbf{\MethodName}.
Extensive experiments demonstrate the effectiveness of \textbf{\DatasetName} and \textbf{\MethodName} in generating expressive, diverse, and realistic human motions.
We also highlight the utility of 2D motion for various downstream applications and its potential for lifting to 3D motion.
The page link is: \url{https://holistic-motion2d.github.io}.
\end{abstract}

\begin{figure}[t]
    \begin{center}
        \centerline{\includegraphics[width=\linewidth]{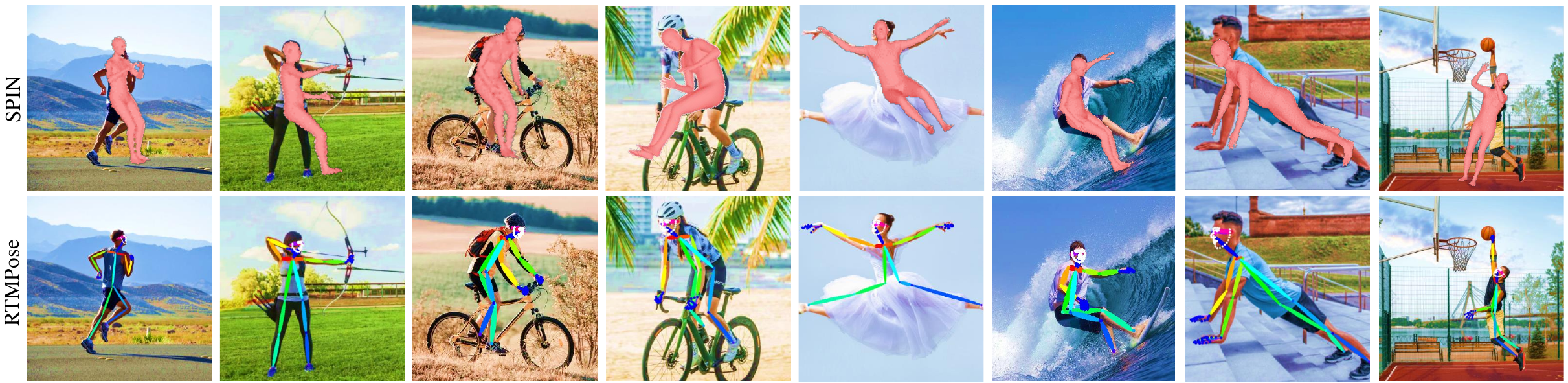}}
        \vspace{-2mm}
        \caption{Comparison of estimating 3D keypoints (SPIN \cite{kolotouros2019learning}) and direct prediction of 2D keypoints (RTMPose~\cite{jiang2023rtmpose}) from images. The precision of 2D keypoints demonstrates robustness to variations in viewpoint.}
        \label{fig:2D-3D}
    \end{center}
\end{figure}


\section{Introduction}
\label{sec:introduction}

Text-driven human motion generation~\cite{chen2023executing,zhang2023t2m,tevet2022human,guo2022generating} is garnering heightened interest due to its potential applications in robotic control, video gaming, and VR/AR. Given a textual prompt, these methods generate motion sequences, which are typically parameterized using the SMPL model~\cite{pavlakos2019expressive}. 
As humans perform a variety of daily activities that require coordinated whole-body movements, including the hands, face, and skeleton, the development of models capable of producing diverse whole-body human movements represents a crucial advance towards comprehensive human motion modeling.

Existing models for generating human motions primarily in 3D space, utilizing either diffusion-based or GPT-based frameworks. While these models demonstrate strong capabilities in producing high-quality motions, 
they are constrained by the limitations of their training datasets, which are typically small, lack diversity, and often include only body skeleton motions.
For example, the widely-used HumanML3D \cite{guo2022generating} dataset, with approximately 14,000 clips, is starkly modest compared to the billions of images~\cite{krizhevsky2012imagenet,zhu2023genimage} used in foundational image generation models~\cite{rombach2022high, ramesh2022hierarchical,saharia2022photorealistic,esser2021taming} and lacks face or hand motion data. 
The scarcity of diverse whole-body 3D human motion data is primarily due to the challenges associated with its acquisition. 
high-precision motion capture (MoCap) systems, capable of capturing detailed human motions, are prohibitively expensive, particularly for whole-body capturing, and are confined to indoor scenarios, limiting scalability.
As shown in Figure~\ref{fig:2D-3D}, 
one alternative involves estimating 3D human motions from monocular videos and refining them via optimization techniques. However, reliably recovering whole-body 3D motion across diverse real-world scenarios remains a formidable challenge, as existing models often robustness and precision.

\textit{Why not model whole-body human motion in 2D space, especially since humans primarily perceive each other's movements in this way?}
2D human motion modeling offers unique advantages over its 3D counterpart, primarily due to the ease of whole-body motion collection. 
Unlike the often ill-posed problem of 3D motion estimation, estimating 2D human poses is markedly more reliable and simpler, especially for whole-body poses (as depicted in Figure~\ref{fig:2D-3D}). Moreover, many applications, particularly in human video generation, require only 2D motion inputs; these models utilize 2D whole-body poses to synthesize high-quality human videos. Further, effective 2D whole-body motion videos serve as strong priors for generating 3D whole-body motion~\cite{lin2023one,cai2024smpler}, similar to how 3D object generation~\cite{poole2022dreamfusion,wang2024prolificdreamer} leverages the prior of Stable Diffusion models~\cite{rombach2022high,podell2023sdxl} . 

In this paper, we propose a novel approach by generating whole-body human motions in 2D space, which is more accessible than 3D motion generation thanks to robust 2D pose estimators.
\textit{Such a novel 2D setting emerges as a valuable supplement to existing 3D motion generation methods.}
To catalyze progress in this new direction, we introduce a large-scale dataset, \textbf{\DatasetName} in Figure~\ref{fig:teaser}. It comprises 1M holistic human sequences, each with precise 2D whole-body (or partial) keypoints and text annotation. Specifically, we develop:

\begin{table}[t]
\small
\centering
\setlength{\tabcolsep}{1.4mm}

\begin{tabular}{lcccccccc}
\toprule
\multirow{2}{*}{Dataset} & \multicolumn{3}{c}{Motion Annotation} & \multicolumn{2}{c}{Text Annotation} & \multicolumn{2}{c}{Scene} & \multirow{2}{*}{\begin{tabular}[c]{@{}c@{}}RGB\\    Image\end{tabular}} \\ 
                         & Clip       & Holistic    & Hours    & Motion          & Holistic        & Indoor      & Outdoor     &                                                                         \\ \midrule
KIT-ML~\cite{plappert2016kit}                   & 3,911       & No            & 11.2     & 6,278            & No                & Yes         & No          & No                                                                      \\
AMASS~\cite{mahmood2019amass}                    & 11,265      & No            & 40.0     & 0               & No                & Yes         & No          & No                                                                      \\
BABEL~\cite{punnakkal2021babel}                    & 13,220      & No            & 43.5     & 91,408           & No                & Yes         & No          & No                                                                      \\
HumanAct12~\cite{guo2020action2motion}               & 1,191          & No            & 6.0        & 1,191               & No                & Yes         & No          & No                                                                      \\
HumanML3D~\cite{guo2022generating}                & 14,616      & No            & 28.6     & 44,970           & No                & Yes         & No          & No                                                                      \\
Motion-X~\cite{lin2024motion}                 & 95,642      & No           & 127.1    & 95,642           & Yes               & Yes         & Yes         & Yes                                                                     \\
Holistic-Motion2D (ours)                     &  1,002,463   & Yes           &      1,614.5   & 1,002,463         & Yes               & Yes         & Yes         & Yes                                                                     \\ \bottomrule
\end{tabular}
\captionsetup{skip=2pt}
\caption{Comparison between our proposed Holistic-Motion2D and existing text-motion datasets. The video quantity of our Holistic-Motion2D is \textbf{10$\times$} larger than the previously largest 3D motion dataset, \textit{i.e.}, Motion-X.}
\label{tab: compared to other dataset}
\end{table}

\paragraph{A large-scale 2D whole-body human motion dataset:} 
We develop a large-scale 2D holistic motion dataset {\DatasetName} leveraging existing large-scale video datasets and designing an automated annotation pipeline for whole-body human motion. 
As demonstrated in Table~\ref{tab: compared to other dataset}, {\DatasetName} excels beyond prior 3D motion datasets in both the volume of video clips and the diversity of human motion scenarios covered, coupled with its holistic motion annotations and descriptive texts.

\paragraph{A powerful 2D whole-body human motion generation baseline model:}
We introduce the 2D whole-body motion generation model {\MethodName}, tailored for 2D whole-body human motion synthesis. 
This model incorporates two novel designs to enhance the quality of generated motions: the \textit{Part-aware Attention for Motion Variational Auto-Encoder (PA-VAE)} and \textit{Confidence-Aware Generation (CAG)}. The PA-VAE module integrates a part-aware spatio-temporal attention mechanism within the motion VAE framework, enabling it to more accurately model fine-grained whole-body movements, such as hand movements. Meanwhile, CAG focuses on prioritizing high-confidence keypoints and accurately inferring positions of occluded parts, thereby minimizing the impact of unreliable pseudo labels. Together, these methodologies equip our model with robust capabilities to handle noisy real-world data, significantly improving 2D motion synthesis accuracy and reliability.

\paragraph{A scalable 2D whole-body human motion evaluation model:} 
Accurately evaluating the quality and semantic fidelity of 2D whole-body motions is important. To address this, we introduce MoLIP, an evaluation model pre-trained in a text-motion contrastive learning manner, which is tailored to measure semantic matching between 2D whole-body motions and texts. Utilizing MoLIP to assess text-motion similarity offers a reliable method for evaluating the semantic fidelity of generated motions, ensuring that they are not only technically accurate but also contextually appropriate.


Comprehensive experiments highlight the importance of our Holistic-Motion2D dataset for generating diverse and realistic 2D whole-body motions. Additionally, we demonstrate several downstream applications of generated 2D human motion, showcasing the broad real-world applicability of our proposed 2D motion generation method.
In summary, our contributions are outlined as follows:
\begin{itemize}[leftmargin=*]
\setlength{\leftmargin}{0pt}
    \item For the first time, we propose the text-driven 2D whole-body motion generation task and construct a \textit{million-level} dataset \textbf{\DatasetName} with high-quality motion and text annotations.  
    \item We develop a 2D whole-body motion generation model \textbf{\MethodName} with two innovative designs, \textit{i.e}, \textit{whole-body Part-Aware VAE} and \textit{Confidence-Aware Generation} to improve 2D motion modeling. 
    \item  We provide a scalable 2D text-motion-aligned model \textbf{MoLIP} for evaluating the semantic fidelity of generated whole-body motions, serving to enclose the semantic gap of text-motion pairs.
\end{itemize}
\section{Related Work}
\label{sec:related_work}
\paragraph{Human Motion Datasets.} 
Benchmarks annotated with sequential human motions and texts pave the way for the development of motion synthesis tasks. Drawing from the marker-based and markless motion capture systems, 3D human motion datasets~\cite{plappert2016kit,mahmood2019amass,guo2022generating,lin2024motion} gain thriving progress. KIT Motion-Language Dataset~\cite{plappert2016kit} annotates sequence-level description for multi-modality motion generation. HumanML3D~\cite{guo2022generating} dataset, derived from the AMASS~\cite{mahmood2019amass} and HumanAct12~\cite{guo2020action2motion} datasets, provides more textual annotation and diverse activities, \textit{e.g.}, sports, acrobatics, and arts. In a departure from them, Motion-X~\cite{lin2024motion} proposes a well-structured annotation protocol and first acquires comprehensive fine-grained 3D whole-body motion dataset from massive scenes. Due to its acquisition, existing text-driven 3D motion datasets exhibit inadequate volume and diversity, resulting in limited scalability for generalized motion synthesis. Towards these issues, we develop a \textbf{Holistic-Motion2D} benchmark in 2D space, featuring diverse scenarios and expanded scale, surpassing the current largest Motion-X.

\paragraph{Text-driven Human Motion Generation.} 
Text-driven human motion generation involves translating descriptive text into corresponding human motion sequences, is gathering increased attention. Previous researches~\cite{guo2022generating,ghosh2021synthesis,ahuja2019language2pose,tevet2022motionclip,zhang2023t2m} focus on modeling a joint latent space for motion and text alignment. MotionCLIP~\cite{tevet2022motionclip} improves the auto-encoder’s generalization by aligning the shared space with the expressive CLIP~\cite{radford2021learning} embedding space. T2M-GPT~\cite{zhang2023t2m} formulates the motion generation as the next-index prediction task by mapping the motion to a sequence of discrete indices. Recent advancements have witnessed diffusion-based generative models renowned for their leading benchmarks in text-to-motion task. MotionDiffuse~\cite{zhang2024motiondiffuse}, MDM~\cite{tevet2022human}, and DiffGesture \cite{zhu2023taming} are pioneering efforts in applying diffusion model into motion generation field. MLD~\cite{chen2023executing} develops a motion latent-based diffusion model to synthesize plausible and diverse human motions. \cite{lu2023humantomato} first proposes text-driven whole-body motion generation task and advance GPT-like model to generate fine-grained motions. However, these 3D-focused methods are not well-adapted for generating 2D motions. To bridge this gap, our Tender is tailored for synthesising 2D whole-body motions with enriched fidelity. 
\begin{figure}[t]
    \begin{center}
        \centerline{\includegraphics[width=\textwidth]{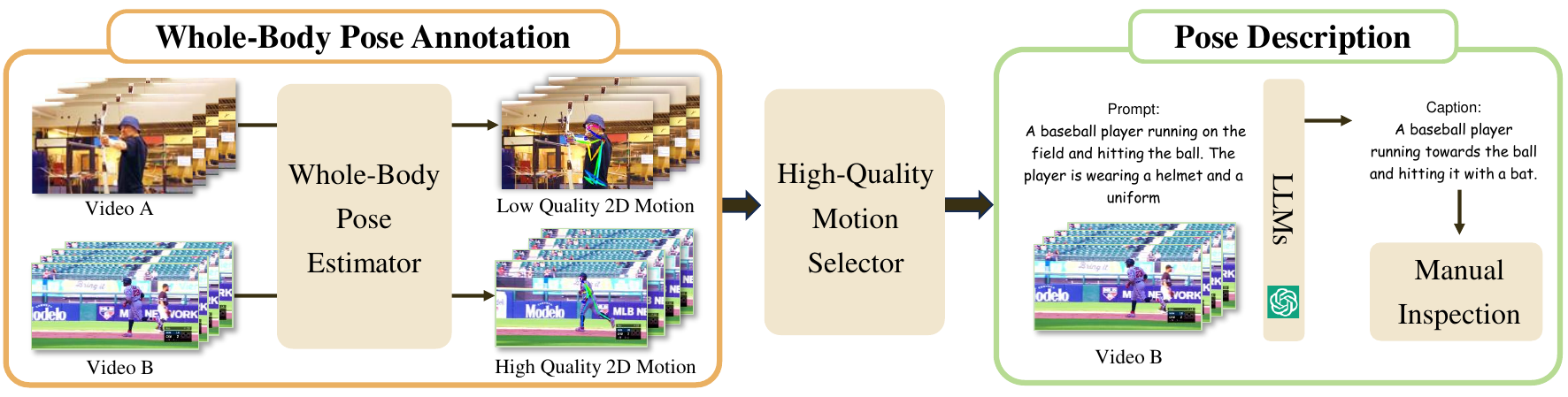}}
        \vspace{-1mm}
        \caption{Overview of the keypoints and pose descriptions annotation pipeline of 2D whole-body motions.}
        \label{fig:pipeline}
    \end{center}
\end{figure}

\section{Holistic-Motion2D: Large-scale 2D Whole-body Motion Dataset}
As shown in Figure~\ref{fig:pipeline}, we illustrate the overall data collection pipeline including holistic 2D motion and caption generation, which includes the following key steps: 1) gathering large volumes of videos, 2) annotating 2D whole-body keypoints and confidence scores, 3) filtering high-quality motion sequences, 4) designing text prompts via Large Language Model (LLM), 5) generating descriptive captions for sequence-level movements, 6) executing the manual inspection. 

\paragraph{Data Source.}
Given contextual dependence of human motions, comprehensive dataset collection across various environments and activities is essential. 2D motions amass from 1M standard videos, including action recognition datasets (UCF101~\cite{soomro2012ucf101}, Kinetics-400~\cite{kay2017kinetics}, Kinetics-700~\cite{kay2017kinetics}, Sth-v2~\cite{goyal2017something}), facial expression datasets (CAER~\cite{lee2019context}, DFEW~\cite{jiang2020dfew}), the video understanding dataset (InternVid~\cite{wang2023internvid}), the upper body dataset (UBody~\cite{lin2023one}), and the 3D motion dataset (IDEA400~\cite{lin2024motion}). Our Holistic-Motion2D encompasses rich scenes, ranging from professional sports (\textit{e.g.}, \textit{playing tennis}, \textit{skiing}) and general daily actions (\textit{e.g.}, \textit{haircut}, \textit{brushing teeth}) to complex human-scene interactions (\textit{e.g.}, \textit{lying down}, \textit{wall push-ups}), capturing diverse environments such as indoor/outdoor landscapes, and dynamic action scenes. Compared with MotionX~\cite{lin2024motion} and HumanML3D~\cite{guo2022generating}, Holistic-Motion2D showcases more elaborate actions, longer motion sequences, and enhanced diversity. 

\paragraph{2D Whole-body Motion Annotations.} Developing an efficient annotation pipeline for high-fidelity 2D whole-body motions enables to capture potential movements from the collected videos. Specifically, we utilize off-the-shelf RTMpose~\cite{jiang2023rtmpose} tool to annotate the 2D whole-body motion $\textbf{K}_p\in\mathcal{R}^{133\times2}$ with confidence scores \textbf{$\textbf{K}_c\in\mathcal{R}^{133\times1}$}, adhering to the COCO-Wholebody~\cite{jin2020whole} format. Compared to MotionX's elaborate keypoint annotation workflow, time-consuming temporal smoothing, and multiple models to handle keypoints of the body regions, our method with streamlined process and unified model, exhibits exceptional scalability and increased friendliness towards large-scale video-level datasets. Please refer to Section \ref{sec:additional_experimental_setting} for selecting high-quality motions.

\paragraph{2D Whole-body Motion Descriptions.} To generate fine-grained motion descriptions of the body, hand and face parts, we develop a comprehensive annotation pipeline as follows: (1) \textit{Coarse-Grained Motion Description Extraction}. We employ the VideoChat2~\cite{li2023mvbench} tool to provide a coarse-grained annotation for our collected videos. By designing prompts, we can use this model to convert rich motion information into text. Since VideoChat2 is not customized for human motion analysis, the resulting captions inevitably contain motion-irrelevant information (environmental context, character clothing) and repetitive expressions, which interfere with model training. Therefore, the extracted captions need further refinement. (2) \textit{Motion Caption Refinement}. Based on coarse-grained motion captions, we perform large-scale reprocessing using ChatGPT. Specifically, we incorporate coarse-grained annotations and designed rules into a dialogue QA, instructing GPT to retain only the motion content itself. Comparing the resulting outputs, we found that GPT effectively cleaned the annotations, significantly improving text quality. Compared to strategies in other datasets for processing textual annotations, our approach, with its streamlined process and consistent model structure, demonstrates exceptional scalability and higher user-friendliness when handling large-scale video-level datasets.

\section{Tender: 2D Whole-body Motion Generation Model}
\label{Method}
\subsection{Problem Formulation}
In this part, we clarify notations and set up the novel benchmarks of 2D text-driven whole-body motion generation. Given the textual description $c$, the task is to synthesize a vivid 2D motion $x^{1:N}$ of frames $N$, which is a sequence of 2D human poses represented by whole-body keypoints $x^i\in\mathcal{R}^{133\times2}$. We formulate the 2D text-driven whole-body motion generation task as:
\begin{equation}
\Theta^{\star}=\operatorname{argmax}_{\Theta} \mathcal{P}(x^{1:N} \mid c, \Theta)
\end{equation}
where $\Theta^{\star}$ denotes optimal model parameters. $\mathcal{P}$ is the motion distribution and $c$ is the text description.

\subsection{Part-Aware Variational Auto-Encoder (PA-VAE)}
\label{Part-Aware Variational Auto-Encoder}
VAEs have demonstrated efficacy in capturing the inherent dynamics and probabilistic nature of human movements. As previous studies~\cite{zhong2023attt2m} have highlighted, the articulated whole-body structures exhibit complicate spatio-temporal relationship. Therefore, the dynamics within individual semantic parts aggregate to dictate the entire human. Previous VAE-based methods overlook spatial modeling, trapped in dilemma when encountering 2D whole-body motion generation tasks. To learn whole-body spatio-temporal characteristics, we propose a \textbf{P}art-\textbf{A}ware \textbf{V}ariational \textbf{A}uto-\textbf{E}ncoder (PA-VAE) to improve expressive ability of latent space and generate contextually coherent whole-body motions.

\begin{figure}[t]
    \begin{center}
        \centerline{\includegraphics[width=\linewidth]{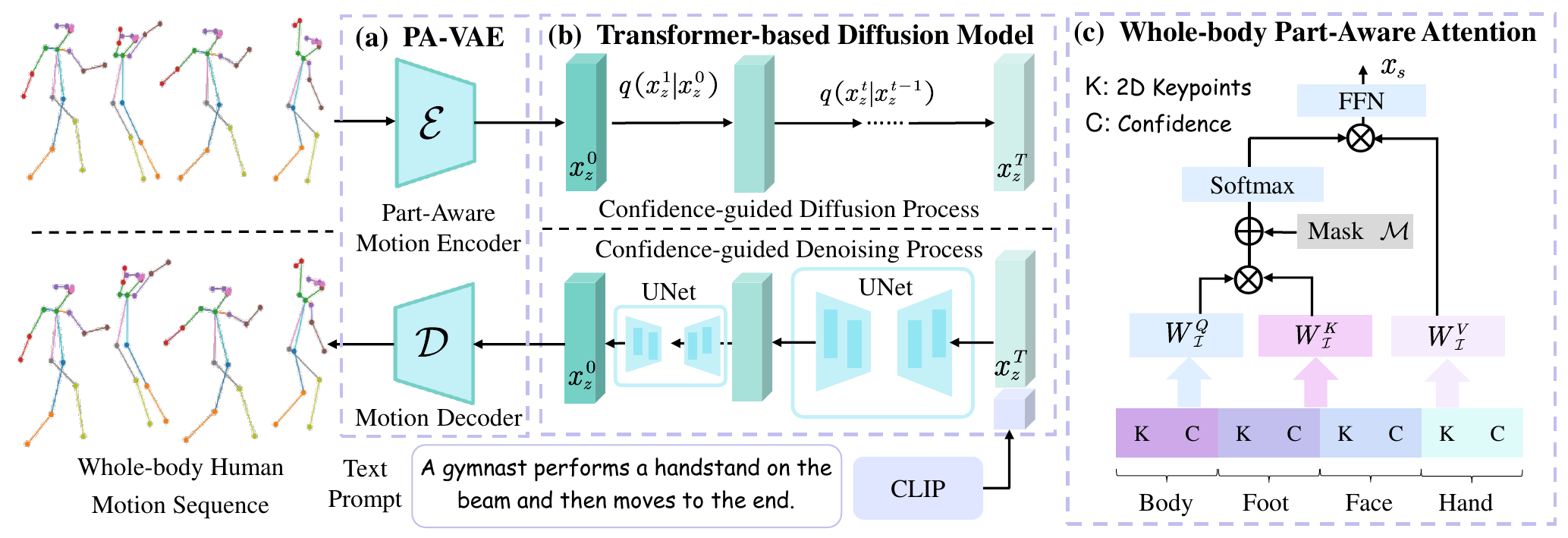}}
        \vspace{-2mm}
        \caption{\textbf{Overview of our Tender framework}. (a) PA-VAE to embed whole-body part-aware spatio-temporal features into a latent space. (b) The diffusion model to generate realistic whole-body motions conditioned on texts. (c) Whole-body Part-Aware Attention to model spatial relations of different parts with CAG mechanism.}
        \label{fig:Framework}
    \end{center}
\end{figure}

As shown in Figure~\ref{fig:Framework}, the PA-VAE module consists of a transformer-based whole-body encoder $\mathcal{E}$ and decoder $\mathcal{D}$. To further enhance the latent representation, we leverage two transformers $\mathcal{E}$ and $\mathcal{D}$ with long skip connections~\cite{ronneberger2015u}. The motion encoder $\mathcal{E}$ takes learnable distribution tokens and motion sequence $x^{1:L}$ of arbitrary length $L$ as input. Then, the encoder $\mathcal{E}$ based on whole-body part-aware attention extracts spatio-temporal representations with high informative density. Specifically, we divide the human body with 133 keypoints into four parts: $\{$Body, Foot, Face, Hand$\}$, each containing its own set of keypoints. Such a structural partition facilitates more focused \textit{hand swing}, \textit{facial expression}, and \textit{foot movements}. We develop a spatial transformer, the heart of which is a whole-body part-aware attention mechanism. Given raw keypoints $x^{1:L}_{\mathcal{I}}\in\mathcal{R}^{L\times P_{\mathcal{I}}C_{\mathcal{I}}}$, $\mathcal{I}$ represents the anatomical divisions $\{\operatorname{Body}, \operatorname{Foot}, \operatorname{Face}, \operatorname{Hand}\}$, $P_{\mathcal{I}}$ and $C_{\mathcal{I}}$ denote the number of keypoints and input dimension for each specific part index $\mathcal{I}$. We embed all motion tokens through linear mapping:
\begin{equation}
Q_{\mathcal{I}}=W^Q_{\mathcal{I}} x^{1:L}_{\mathcal{I}}\in\mathcal{R}^{L\times C}, K_{\mathcal{I}}=W^K_{\mathcal{I}} x^{1:L}_{\mathcal{I}}\in\mathcal{R}^{L\times C}, V_{\mathcal{I}}=W^V_{\mathcal{I}}x^{1:L}_{\mathcal{I}}\in\mathcal{R}^{L\times C}
\end{equation}
where $W^Q_{\mathcal{I}}$, $W^K_{\mathcal{I}}$, and $W^V_{\mathcal{I}}$ are projection matrices. Upon channel-wisely concatenating the projected motion tokens, we establish the final $Q$, $K$, $V$. Moreover, an adjacency mask $\mathcal{M}\!=\!\{m_{ij}\}\!\in\!\mathcal{R}^{133\times133}$ is specified, aligned with body part divisions. $m_{ij}$ is designated as zero if keypoints $i$ and $j$ belong to the same part, otherwise $-\infty$. Such an explicit partition protocol refines the allocation of attention and captures interaction of body parts. The part-aware attention is formulated as:
\begin{equation}
\resizebox{0.4\hsize}{!}{$
x_{s}=\operatorname{FFN}\left[\operatorname{softmax}\left(\frac{Q K^T \oplus \mathcal{M}}{\sqrt{C}}\right) V\right],$}
\end{equation}
The spatial-enhanced motion representation $x_{s}$ is fed into several skip-connected transformer layers to maintain temporal coherence and model long-term dependencies across whole-body motions, resulting in the motion temporal feature $x_{t}$. The embedded tokens serve as Gaussian mean $\mu$ and deviation $\sigma$ of the motion latent space $\mathcal{Z}$ to re-parameterize latent $x_{z}\in\mathcal{R}^{n\times d}$. Finally, the transformer-based decoder $\mathcal{D}$ takes $x_{z}$ as zero-padding motion tokens as queries and latent $x_{z}$ as keys to reconstruct the whole-body motion sequence with cross attention mechanism. 

\subsection{Confidence-Aware Generation (CAG)}
\label{Confidence-Infused Motion Latent Modeling}
In addition to limited spatio-temporal modeling prowess, current VAEs also struggle with pronounced occlusions of whole-body parts, a frequent occurrence in dynamic activities. Further, the limited understanding of human motion variability by the VAE restricts its ability to generalize when confronted with unseen motions. To this issue, we develop a \textbf{C}onfidence-\textbf{A}ware \textbf{G}eneration (CAG) mechanism to prioritize high-confidence keypoints and minimize adverse effects of unreliable ones. 

We concatenate confidence scores with 2D whole-body keypoints as inputs to PA-VAE, obtaining confidence-infused motion latents. Incorporating the CAG mechanism, our Tender gains improved perceptual abilities of human motions by modeling the distribution of confidence. We further leverage the diffusion model to jointly capture the spatial layouts and confidence distributions of 2D keypoints.

Diffusion models~\cite{song2020denoising,song2020improved,rombach2022high,dhariwal2021diffusion}, grounded in principles of stochastic diffusion process, have showcased commendable generative prowess in human motion generation. To improve the quality and flexibility of whole-body motions, we perform a transformer-based denoising model, $G_\theta$, equipped with long-skip connections~\cite{ronneberger2015u} on abstract and generalized motion latents $x_{z}\in\mathcal{R}^{n\times d}$, which encodes confidence score and undergoes diffusion modeled as a Markov noising process: 
\begin{equation}
\label{diffusion}
\resizebox{0.4\hsize}{!}{$q\left(x_{z}^{t} \mid x_{z}^{t-1}\right)=\mathcal{N}\left(\sqrt{\alpha_t} x_{z}^{t-1},\left(1-\alpha_t\right) I\right),$}
\end{equation} 
where $\alpha_t\in(0,1)$ are constant hyper-parameters for sampling and $x_{z}^{t-1}$ denotes the motion latent at noising step $t$. The above forwarding Markov chain is reversed to learn the original joint distributions of motion and confidence. Instead of predicting noise, we follow MDM~\cite{tevet2022human} to predict the signal itself:
\begin{equation}
\resizebox{0.6\hsize}{!}{$
\mathcal{L}_{\text {simple}}=E_{x_{z}^{0} \sim q\left(x_{z}^{0} \mid \phi_\theta(c)\right), t \sim[1, T]}\left[\left\|x_{z}^{0}-G_\theta\left(x_{z}^{t}, t, \phi_\theta(c)\right)\right\|_2^2\right],$}
\end{equation}
$\phi_\theta(c)$ represents the CLIP text encoder, which is freezed during the parameter optimization. In our work, we execute text-driven 2D whole-body motion synthesis by conditioning on CLIP in a classifier-free guidance manner, which provides a trade-off diversity and fidelity by interpolating or potentially extrapolating both the conditioned and the unconditioned distributions:
\begin{equation}
G_\theta^s\left(x_{z}^{t}, t, c\right)=s G_\theta\left(x_{z}^{t}, t, c\right)+(1-s) G_\theta\left(x_{z}^{t}, t, \varnothing\right)
\end{equation}
where $s>1$ is the guidance scale. Subsequent to the interactive reverse iteration of the conditional denoising, the decoder $\mathcal{D}$ reconstructs the motion from the predicted $x_{z}^{0}$.

\subsection{MoLIP: 2D Text to Whole-body Motion Retrieval Model}
Previous widely-used evaluation model for 3D human motions is trained on HumanML3D \cite{guo2022generating}, which is inapplicable to our 2D-focused task.
In this case, we develop a more generalizable retrieval model \textbf{MoLIP} for evaluating the quality of the generated 2D whole-body motions.
Following HumanML3D \cite{guo2022generating}, we train a CLIP-style \cite{radford2021learning} model, including a motion encoder $\mathcal{E}_x$ and a text encoder $\mathcal{E}_c$, to match 2D text-motion pairs.
We aim to learn structurally-aligned features of the motion encoder $\mathcal{E}_x$ and text encoder $\mathcal{E}_c$ in the latent space. Given the motion and text as $x$ and $c$, the training loss is defined as: 
\begin{equation}
\resizebox{0.75\hsize}{!}{$\mathcal{L}_{\text{ret}}=0.5\cdot \left[\text{CE}\left(\mathcal{E}_x\left(\mathcal{T}\left[x\right]\right) \cdot \mathcal{E}_c\left(\mathcal{T}\left[c\right]\right)^T, \mathcal{Y}\right)+\text{CE}\left(\mathcal{E}_c\left(\mathcal{T}\left[c\right]\right) \cdot \mathcal{E}_x\left(\mathcal{T}\left[x\right]\right)^T, \mathcal{Y}\right)\right],$}
\label{equ:loss_retrieval}
\end{equation}
where $\text{CE}(\cdot)$ is the cross-entropy loss, $\mathcal{T}[\cdot]$ denotes $\mathcal{L}_2$ normalization. $\mathcal{Y}=[1,2,\cdots,B]$ is the ground truth labels with batch size $N$. $x \in \mathcal{R}^{B \times N \times 133 \times 2}$ and $N$ is the length of the input motion.

\begin{figure}[t]
\centering
    \includegraphics[width=\linewidth]{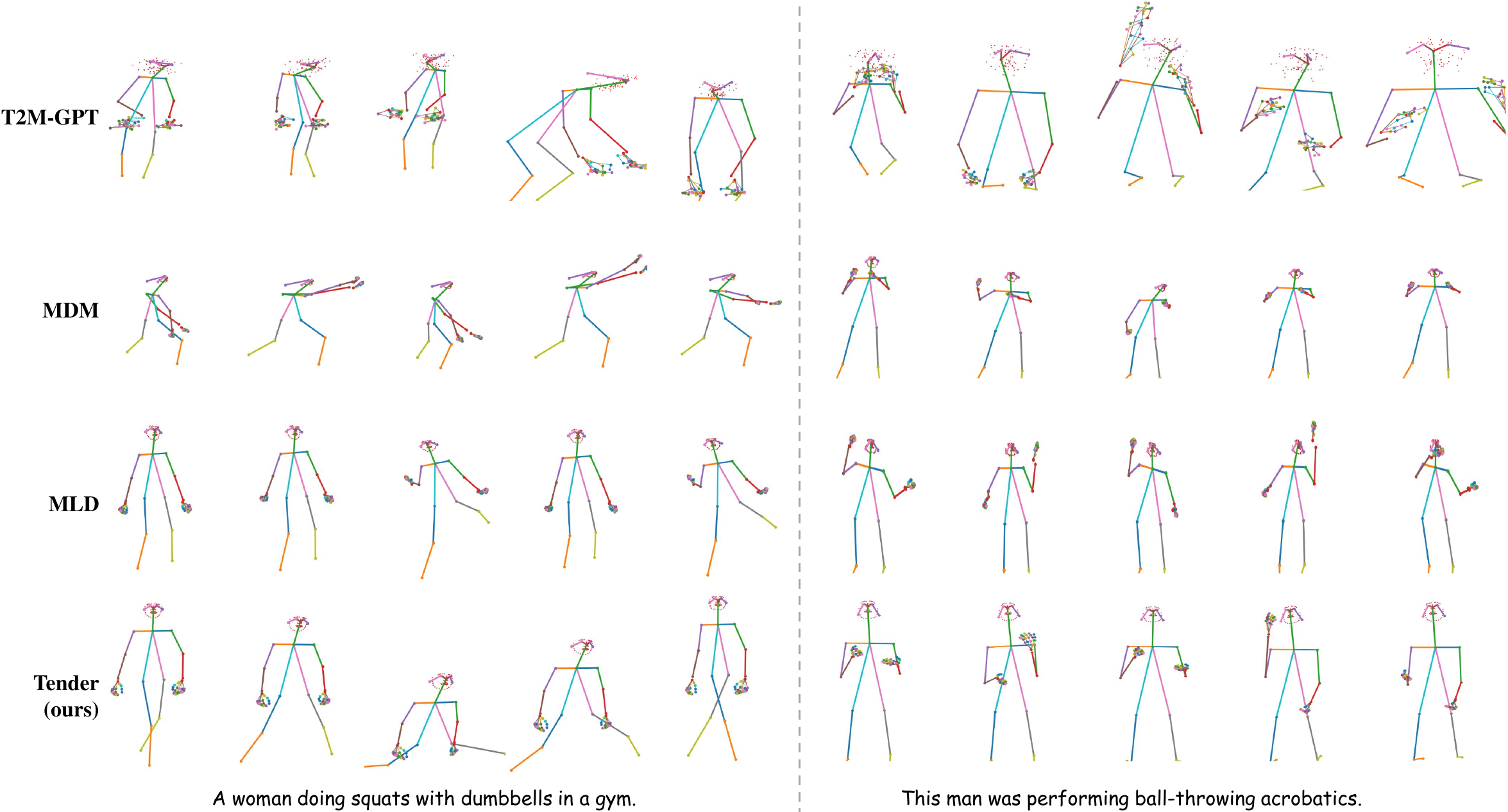}
    \vspace{-5mm}
    \caption{{\bf Qualitative results of our Tender compared with previous SOTA methods.} Our Tender generates clearly more vivid human motions and preserves the fidelity, together with superior temporal consistency.}
    \label{fig:qualitative_comparison}
\end{figure}


\begin{table}[t]
\small
\centering
\setlength{\tabcolsep}{2.6mm}
{
\begin{tabular}{lccccccc}
\toprule
\multirow{2}{*}{Method} & \multicolumn{3}{c}{R Precision} & \multirow{2}{*}{FID$\downarrow$} & \multirow{2}{*}{MM Dist$\downarrow$} & \multirow{2}{*}{Diversity$\uparrow$} & \multirow{2}{*}{MModality$\uparrow$} \\ 
& Top1$\uparrow$ & Top2$\uparrow$ & Top3$\uparrow$ \\
\midrule
Real Motions &  0.6710 & 0.8620  & 0.9381 & 0.0025 & 5.5754 & 39.87 & —\\ 
\midrule
T2M-GPT~\cite{zhang2023t2m} & 0.1434 & 0.2438  & 0.3186 & 27.8674 & 29.9663 & 36.73 & 35.59 \\
MDM~\cite{tevet2022human} & 0.3355 & 0.5074 & 0.6109 &  6.8437 & 21.2664 & \textbf{38.64} & 38.18 \\
MLD~\cite{chen2023executing} & 0.3639 & 0.5597 & 0.6676 & 4.1184 & 17.4303 & 37.37 & 36.21 \\
Tender (ours) & \textbf{0.3761} & \textbf{0.5736} & \textbf{0.6902} & \textbf{3.9038} & \textbf{16.8211} & 38.42 & \textbf{38.36} \\ 
\bottomrule
\end{tabular}}
\captionsetup{skip=2pt}
\caption{\textbf{Quantitative results of text-to-motion generation on the test set of the proposed Holistic-Motion2D dataset}. The symbol `$\uparrow$'(`$\downarrow$') indicates that the higher (lower) is better.
``MModality'' denotes MultiModality. 
}
\label{Quantitative results}
\end{table}

\section{Experiment}
\label{sec:experiment}

All our experiments are trained with AdamW~\cite{loshchilov2017decoupled} optimizer with a fix learning rate of 2e-4 without weight decay. Our Tender is trained with 4 NVIDIA 80G A100 GPUs and the batch size on each GPU is set to 32/256 for PA-VAE/diffusion model training. We select three dataset configurations: \textbf{(D1)} The whole high-quality subset; \textbf{(D2)} high-quality data from K400 and K700; \textbf{(D3)} 20\% of the high-quality dataset to establish the benchmark and validate the effectiveness of our Tender.
\subsection{Compared with Other Text-to-motion Methods}
As shown in Table~\ref{Quantitative results}, we compare our Tender with existing text-driven motion generation baselines with \textbf{(D1)}. T2M-GPT~\cite{zhang2023t2m} obtains discrete motion representations with VQVAE~\cite{van2017neural} and generate motion tokens in a GPT-like~\cite{floridi2020gpt} fashion. The discrete representation leads to information loss of nuanced motion, while the autoregressive nature of GPT makes it prone to error accumulation, especially when dealing with low-confidence keypoints in 2D whole-body motions. Compared with the diffusion-based MLD~\cite{chen2023executing} and MDM~\cite{tevet2022human}, our Tender captures fine-grained motion representations and is aware of the occlusion and partial visibility of whole-body parts assisted by the CAG. We achieve the best performance in FID and R-Precision, outperforming other three baseline methods on generation quality and text-motion alignment. Figure~\ref{fig:qualitative_comparison} also indicates our method correctly matches the text prompt while maintaining a rich diversity of generated vivid human motions.

\subsection{Scaling Capability of Generation and Retrieval Models}
\begin{wraptable}[12]{r}{6.cm}
\centering
\small
\setlength{\tabcolsep}{1.6mm}
\scalebox{0.85}{

\begin{tabular}{ccccc}
\toprule
\setlength{\tabcolsep}{1mm}
Data  & Model Size & Top1$\uparrow$ & FID$\downarrow$ & Diversity$\uparrow$ \\
\midrule
20\%  & \multirow{4}[2]{*}{9 layers} & 0.3379 & 7.7581 & 37.54 \\
50\%  &       & 0.3664 & 7.0489 & 37.29 \\
80\%  &       & 0.3697 & 6.5420 & 37.40 \\
100\% &       & 0.3761 & 3.9038 & 38.42 \\
\midrule
20\%  & \multirow{4}[2]{*}{5 layers} & 0.2965 & 22.4394 & 37.39 \\
50\%  &       & 0.3234 & 12.4010 & 37.54 \\
80\%  &       & 0.3277 & 8.9618 & 37.67 \\
100\% &       & 0.3375 & 10.5134 & 37.71 \\
\bottomrule
\end{tabular}%
}
\captionsetup{skip=2pt}
\caption{Scaling-up performance analysis for our proposed Tender model on the (\textbf{D1}) subset.}
\label{Scaling law}
\end{wraptable}

\paragraph{2D Whole-body Motion Generation Model.}
Results in Table~\ref{Scaling law} demonstrate the impact of data volume and model size on performance of 2D whole-body motion generation models. It is observed that 1) increasing the scaling of training data consistently improves performance across multiple metrics. Increased training data enhance the model's generalization capabilities, leading to more accurate generation of motion sequences aligned with textual descriptions and heightened fidelity of whole-body motions. 2) Model with a larger scale performs better at almost any given amount of data than a smaller one. For instance, the larger model provides a robust framework to learn richer motion representations and understand more complex spatial-temporal dynamics of movements.


\begin{wraptable}[8]{r}{4.8cm}
\vspace{-2mm}
\centering
\small
\setlength{\tabcolsep}{2mm}
\scalebox{0.85}{
\begin{tabular}{cccc}
\toprule
Data  & Top1$\uparrow$ & FID$\downarrow$ & Diversity$\uparrow$ \\
\midrule
20\%  & 0.2538 & 0.0165 & 43.29 \\
50\%  & 0.4895 & 0.0075 & 41.80 \\
80\%  & 0.6023 & 0.0099 & 41.96 \\
100\% & 0.6710 & 0.0025 & 39.87 \\
\bottomrule
\end{tabular}%
}
\captionsetup{skip=2pt}
\caption{Scaling-up performance analysis for MoLIP on the (\textbf{D1}) subset.}
\label{tab:scaling_law_t2m_model}%
\end{wraptable}%

\paragraph{2D Text-motion Retrieval Model.}

We further study the effect of data scale to our retrieval model, as shown in Table~\ref{tab:scaling_law_t2m_model}.
We conduct experiments on 20\%, 50\%, 80\%, and 100\% high-quality text-motion pairs.
With the increase of the training data, the performance of our retrieval model improves significantly. 
For example, increasing the data from 20\% to 50\% provides almost \textbf{1x} gain on Top1 and FID, which strongly demonstrates the necessity of scaling up of data used for training a generalizable text-to-motion retrieval model.

\subsection{Effect of Data Source and Quality}
\paragraph{Low-quality Motions.}
In Table~\ref{low quality motion data}, we investigate the effects of low-quality motions for training motion generation model. The methods tested include training solely with high-quality data (\textit{Only high}), augmenting high-quality data by $x$ times (\textit{Argument-$x$}), and a low quality pre-training then high-quality fine-tuning approach (\textit{Fine-tune}). While the \textit{Only-high} setting provides excellent fidelity in motion generation, incorporation of low-quality data offers a promising compromise between quality and diversity of our Tender model. The \textit{Argument-x} experiments reveal interesting trends compared to the \textit{Only-high} setting, particularly in \textit{Argument-4} setups. Large-scale low-quality motions are instrumental in broadening action types and enriching text-to-motion instances. It empowers to generate whole-body motions in varying contexts and learn generalized cross-modal alignment. When evaluated with a retrieval model pre-trained using incorporated confidence (\textbf{highlighted rows}), superior results are witnessed, validating the effectiveness of our text-motion-aligned model MoLIP.

\paragraph{Multi-source Datasets.}
In Table~\ref{Multi-Source Data Integration}, we conduct a series of experiments to investigate the impact of integrating additional data sources, such as 3D and facial data. Compared with the MLD~\cite{chen2023executing} method, our Tender showcases remarkable improvements upon the introduction of facial data, where the proposed PA-VAE provides more granular modeling of facial expressions (\textit{e.g.}, angry, smile, or frown) and facilitates improved natural and expressive whole-body motions. Further, additional 3D motion dataset (IDEA400) a broader variety of motion patterns and dynamics, encourage the proposed Tender to generate complex, realistic, and contextually rich whole-body motions.

\begin{table}[t]
\small
\begin{minipage}[c]{0.48\textwidth}
\centering
\setlength{\tabcolsep}{1.9mm}
\begin{tabular}{lccc}
\toprule
Part-aware VAE & Top1$\uparrow$    & FID$\downarrow$     & Diversity$\uparrow$ \\ \midrule
Naive VAE            & 0.3120 & 6.2446  & 37.78     \\
PA-VAE-3-4     & 0.3159 & 5.7189  & 37.39     \\
PA-VAE-2-4     & \textbf{0.3250} & \textbf{4.4032}  & \textbf{37.86}     \\
PA-VAE-2-3     & 0.3215 & 7.0700  & 37.64     \\
PA-VAE-2-2     & 0.2636 & 6.3866  & 37.81     \\
PA-VAE-1-4     & 0.2821 & 6.4368  & 36.99     \\
PA-VAE-1-2     & 0.2524 & 10.0147 & 37.70     \\ \bottomrule
\end{tabular}
\captionsetup{skip=2pt}
\captionof{table}{Comparison of different whole-body part-aware VAE configurations on the (\textbf{D2}) subset. ``PA-VAE-2-4'' performs best.}
\label{Part-aware VAE}
\end{minipage}
\hspace{0.05\textwidth}
\begin{minipage}[c]{0.44\textwidth}
\small
\setlength{\tabcolsep}{2mm}
\centering
\begin{tabular}{lccc}
\toprule
Method     & Top3$\uparrow$     & FID$\downarrow$    & Diversity$\uparrow$ \\ \midrule
Only High  & 0.6774 & 3.9038 & 38.42     \\
\rowcolor[HTML]{FFF2CC} 
Only High  & 0.6950 & 2.4440 & 41.38     \\
Finetune   & 0.6442 & 8.0525 & 41.56     \\
Argument-1 & 0.6674 & 7.9846 & 40.99     \\
Argument-2 & 0.6995 & 6.5995 & 41.24     \\
Argument-4 & 0.6901 & 4.8562 & 41.52     \\ 
\rowcolor[HTML]{FFF2CC} 
Argument-4 & 0.7077 & 3.7428 & 42.08    \\ \bottomrule
\end{tabular}
\captionsetup{skip=2pt}
\captionof{table}{Impact of low-quality motions on the (\textbf{D1}) subset. Highlighted rows denote the results evaluated by MoLIP trained with confidence.}
\label{low quality motion data}
\end{minipage}
\vspace{-0.3cm}
\end{table}

\begin{table}[t]
\centering
\small
\begin{minipage}[c]{0.48\textwidth}
\centering
\setlength{\tabcolsep}{0.58mm}
\begin{tabular}{cccccc}
\toprule
Velocity & Weight & CAG & Top1$\uparrow$   & FID$\downarrow$     & Diversity$\uparrow$ \\ \midrule
         &          &             & 0.3250 & 4.4032  & 37.86     \\
\checkmark        &          &             & 0.3453 & 6.4038  & 37.16     \\
         & \checkmark        &             & 0.3131 & 8.5602  & 38.03     \\
         &          & \checkmark           & 0.3285 & 3.5675  & 36.13     \\
\checkmark        & \checkmark        &             & 0.3244 & 10.1125 & 37.40     \\
\checkmark        &          & \checkmark           & 0.3531 & 3.6970  & 40.17     \\
\checkmark        & \checkmark        & \checkmark           & 0.3476 & 5.8864  & 38.59     \\ \bottomrule
\end{tabular}
\captionsetup{skip=2pt}
\caption{Ablations of the approaches to utilizing the whole-body confidence score on the (\textbf{D2}) subset.}
\label{confidence}
\end{minipage}
\hspace{0.04\textwidth}
\begin{minipage}[c]{0.44\textwidth}
\small
\setlength{\tabcolsep}{0.8mm}
\centering
\begin{tabular}{lccc}
\toprule
Multi-domain Data & Top1$\uparrow$     & FID$\downarrow$    & Diversity$\uparrow$ \\ \midrule
MLD               & 0.3120 & 6.2446 & 37.78     \\
Ours (whole-body) & 0.3250 & 4.4032 & 37.86     \\ 
MLD (w/ face)     & 0.3299 & 7.2927 & 41.20     \\
Ours (w/ face)    & 0.3443 & 4.1074 & 38.81     \\  
MLD (w/ IDEA400)  & 0.3279 & 7.9774 & 38.28     \\
Ours (w/ IDEA400) & 0.3388 & 3.6376 & 39.17     \\ \bottomrule
\end{tabular}
\captionsetup{skip=2pt}
\caption{Impact of multi-source data integration on 2D whole-body motion generation on the (\textbf{D2}) subset. Our method performs admirably.}
\label{Multi-Source Data Integration}
\end{minipage}
\end{table}

\subsection{Ablation Studies}
\paragraph{Ablation Studies of the PA-VAE module.} In Table~\ref{Part-aware VAE}, we ablate several designs of the proposed PA-VAE module in a controlled setup. The ``Part-VAE-\textit{s}-\textit{p}'' denotes the PA-VAE module with \textit{s}-layer spatial transformer and \textit{p} body parts, while ``VAE'' serves as the baseline model without part-aware spatial attention. Table~\ref{Part-aware VAE} presents compelling evidence that our PA-VAE model, configured with 4 part divisions [body, foot, face, hand] and 2 spatial transformer layers achieves optimal performance. Our PA-VAE achieves a marked improvement over the vanilla VAE across all metrics.

\paragraph{Approaches to Utilizing Confidence.} In Table~\ref{confidence}, our inquiry centers on the utilization of confidence scores to better handle the occlusions and partial visibility of bodies. We then develop two alternative methods: 1) employing confidence scores as keypoint-specific weights to modulate the reconstruction loss in the training of PA-VAE (\textit{weighted}), 2) our proposed CAG method. Notably, the weighted confidence loss results in a decrease across all metrics. It can not optimally preserve motion details and introduce biases where low-confidence keypoints exist. From Table~\ref{confidence}, the CAG improves our Tender's performance in generating motions that are both visually appealing and closely aligned with the textual descriptions, particularly when combined with velocity constraints.

\subsection{Downstream Applications}
\label{Downstream Applications}
\begin{figure}[htbp]
\centering
\includegraphics[width=\linewidth]{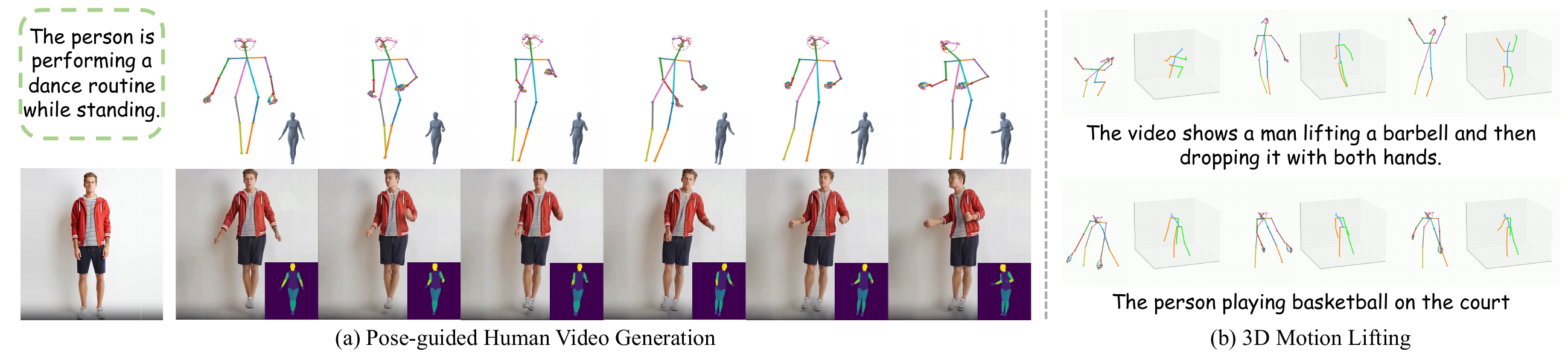}
\vspace{-6mm}
\caption{\textbf{Qualitative results of two practical downstream applications} driven by our proposed Tender method: (a) Pose-guided Human Video Generation. (b) 3D Motion Lifting.}
\label{fig:downstream}
\end{figure}

Recently, conditioned video generation \cite{xu2023magicanimate, zhang2023controlvideo, wang2024customvideo} and 3D pose estimation \cite{zhu2023motionbert} becomes popular.
In Figure~\ref{fig:downstream}, we apply our generated motions to two downstream applications, {\it (a) pose-guided human video generation} and {\it (b) 3D motion lifting}.
Using MagicAnimate~\cite{xu2023magicanimate} to animate the human character with our generated pose sequence and MotionBERT~\cite{zhu2023motionbert} to conduct 3D motion lifting, it unveils visually-compelling results.
From Figure~\ref{fig:downstream}, we find that both animated human-centric videos and lifted 3D human motions are smooth and vivid, demonstrating the practical utility of our method.

\section{Conclusion}
\label{sec:conclusion}
For the first time, we present a pioneering method for generalized human motion generation in 2D space, which addresses limited dataset size and diversity inherent in 3D motion synthesis. Our well-established Holistic-Motion2D dataset and benchmark, enriched with over 1M in-the-wild high-quality whole-body pose $\&$ textual description pairs. Rooted in it, a powerful baseline model Tender-2D is engineered for 2D whole-body motion generation, equipped with innovative whole-body part-aware attention and confidence-aware modeling techniques, which delivers superior performance in motion realism and diversity. Further, we propose a pre-trained text-motion-aligned model for evaluate the semantic fidelity of generated 2D whole-body motions. We also highlights the potential of 2D motion data in downstream applications, \textit{e.g.}, virtual character control and 3D motion lifting.

\appendix



\section{Additional Details for Holistic-Motion2D}
\label{Additional Details for Holistic Motion-2D}
In this section, we delve deeper into the \textbf{Holistic-Motion2D} dataset, offering extended information not fully detailed in the main manuscript due to space limitations. We elaborate on the individual sub-dataset that comprises our comprehensive collection, presenting a thorough statistical analysis of their \textit{scale} and \textit{diversity}. Additionally, we provide essential references for their licensing details and address ethical considerations pertinent to their use and distribution. This supplemental information ensures transparency and facilitates responsible utilization of the dataset in future research endeavors.

\subsection{Detailed descriptions of Sub-dataset.}
\label{Detailed descriptions of Sub-dataset.}
This section outlines the 10 datasets under investigation in our study, as shown in Table \ref{sub-dataset}. It is essential to highlight that these datasets are publicly available within the academic domain, each governed by its unique licensing terms.  In strict adherence to ethical guidelines, we employ these datasets exclusively for non-commercial research purposes. We advocate for interested readers to consult the official websites or research papers associated with each dataset for a deeper understanding of licensing agreements and privacy policies, ensuring the protection of sensitive information. 

\begin{table}[htbp]
\setlength{\tabcolsep}{5.2mm}
\centering
\small
\begin{tabular}{cllcc}
\toprule
\begin{tabular}[c]{@{}l@{}}Data\end{tabular} & \#Clip & \#Frame & Motion     & Text              \\ \midrule
UCF101~\cite{soomro2012ucf101}                                                                                    &   11,387   &    1,692,772   & Whole-Body & Caption           \\
K400~\cite{kay2017kinetics}                                                                                      &   215,476   &   28,792,274    & Whole-Body & Caption           \\
K700~\cite{kay2017kinetics}                                                                                      &   228,508   &    38,116,834   & Whole-Body & Caption           \\
UBody~\cite{lin2023one}                                                                                     &   5,194   &   809,180    & Whole-Body & Caption           \\
InternVid~\cite{wang2023internvid}                                                                                 &   310,260   &    39,285,981   & Whole-Body & Caption           \\
HAA500~\cite{chung2021haa500}                                                                                   &   7,989   &    403,515   & Whole-Body & Caption          \\
IDEA400~\cite{lin2024motion}                                                                                   &   12,025   &    2,129,917   & Whole-Body & Semantic          \\
Sthv2~\cite{goyal2017something}                                                                               &   192,558   &    3,808,216   & Hand       & Semantic          \\
DFEW~\cite{jiang2020dfew}                                                                                      &   15,524   &   1,104,717    & Face       & Semantic          \\
CARE~\cite{lee2019context}                                                                                      &   3,542   &   98,376    & Face       & Semantic          \\
Ours sum                                                                                      &   1,002,463   &    116,241,782   & Whole-Body & Semantic, Caption \\ \bottomrule
\end{tabular}
\captionsetup{skip=2pt}
\caption{Statistics of the sub-datasets in our proposed \textbf{Holistic-Motion2D} dataset. ``Semantic'' and ``Caption'' denote the \textit{semantic labels} (e.g., action categories and expression types) and \textit{textual descriptions}.}
\label{sub-dataset}
\end{table}

\begin{figure}[t]
    \begin{center}
        \centerline{\includegraphics[width=\linewidth]{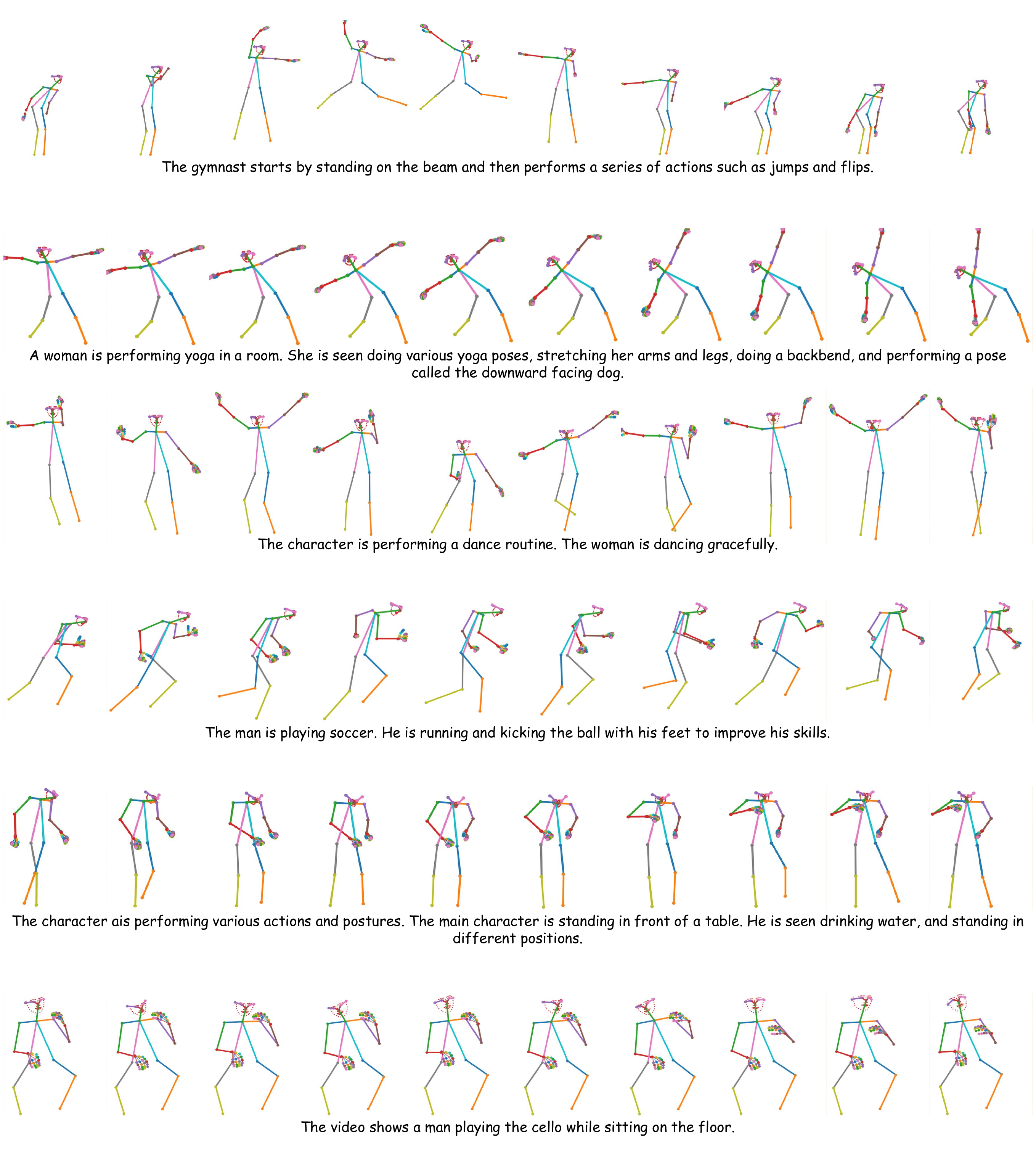}}
        \vspace{-2mm}
        \caption{Visualization results of \textbf{\DatasetName} in multiple visual scenarios.} 
        \label{fig:visual_results}
    \end{center}
\end{figure}

\begin{figure}[htbp]
    \begin{center}
        \centerline{\includegraphics[width=0.97\linewidth]{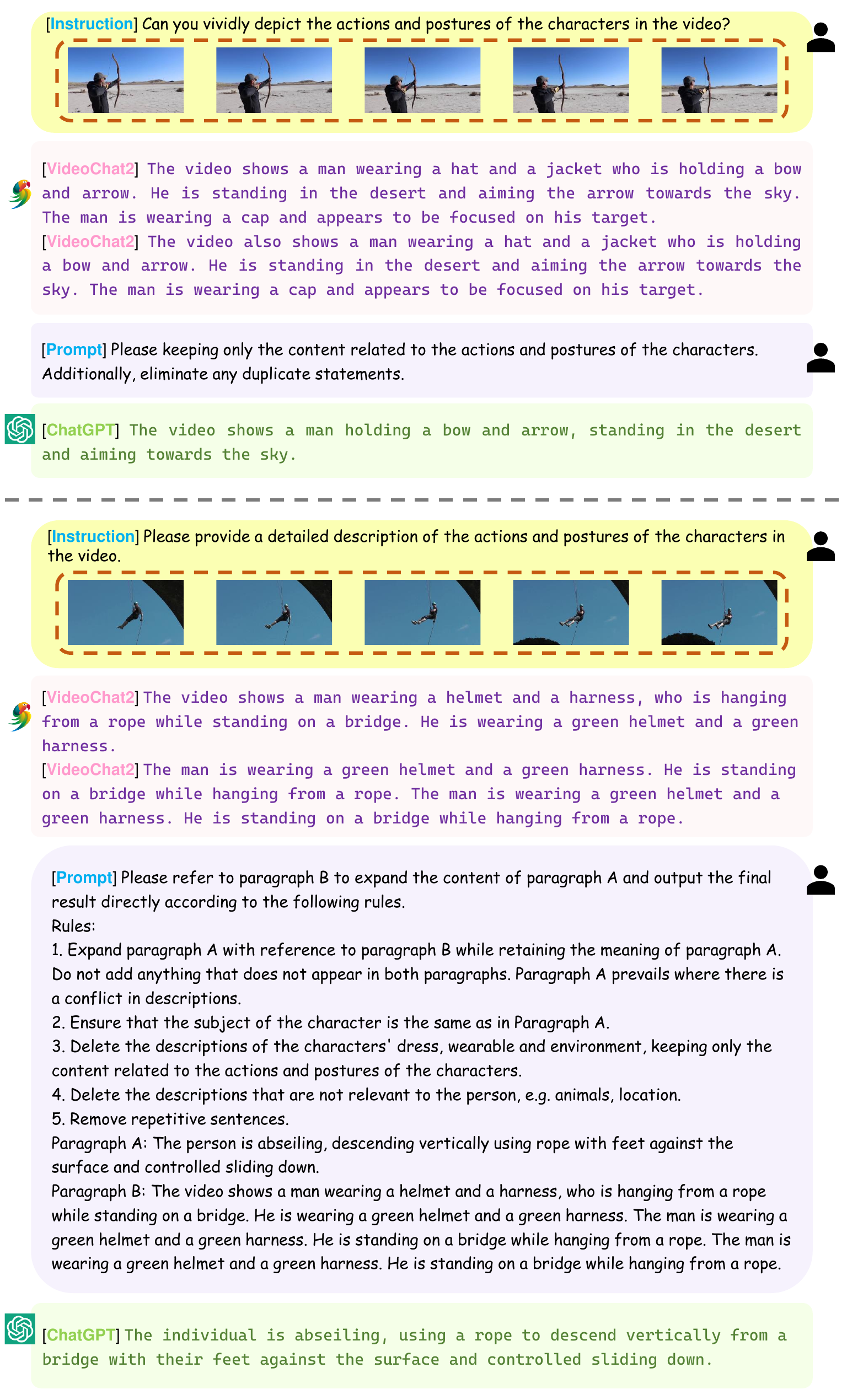}}
        \vspace{-2mm}
        \caption{{\bf Visualization results for annotating the caption of \textbf{\DatasetName}} in multiple visual scenarios.} 
        \label{fig:caption_visualization}
    \end{center}
\end{figure}

\begin{table}[t]
\centering
\small
\setlength{\tabcolsep}{1.5mm}
\begin{tabular}{ccccclc}
\toprule
\multirow{2}{*}{Data} & \multicolumn{3}{c}{R precision}                  & \multirow{2}{*}{FID$\downarrow$} & \multirow{2}{*}{MM Dist$\downarrow$} & \multirow{2}{*}{Diversity$\uparrow$}  \\
                      & Top1$\uparrow$ & Top2$\uparrow$ & Top3$\uparrow$ &                                  &                                      &                                                                        \\ \midrule
20\%                  & $0.2538^{\pm0.0005}$         & $0.3982^{\pm0.0005}$        & $0.5018^{\pm0.0006}$         & $0.0165^{\pm0.0011}$                           & $19.7802^{\pm0.0074}$                              & $43.29^{\pm0.3216}$                                                                    \\
50\%                  & $0.4895^{\pm0.0007}$         & $0.6527^{\pm0.0006}$        & $0.7352^{\pm0.0005}$         & $0.0086^{\pm0.0005}$                           & $13.0567^{\pm0.0058}$                              & $41.80^{\pm0.2908}$                                                                  \\
80\%                  & $0.6023^{\pm0.0008}$         & $0.7809^{\pm0.0005}$         & $0.8606^{\pm0.0004}$         & $0.0099^{\pm0.0008}$                           & $7.2429^{\pm0.0055}$                               & $41.96^{\pm0.2619}$                                                                   \\
100\%                 & $0.6710^{\pm0.0005}$         & $0.8620^{\pm0.0005}$         & $0.9381^{\pm0.0003}$         & $0.0025^{\pm0.0003}$                           & $5.5754^{\pm0.0029}$                               & $39.87^{\pm0.1759}$                                                                  \\ \bottomrule
\end{tabular}
\captionsetup{skip=2pt}
\caption{Scaling-up performance analysis for MoLIP on the (D1) subset across all performance metrics. Experiments underscores our pre-trained MoLIP's robust scalability in aligning motion generation with texts. }
\label{Scaling law-more}
\end{table}

\begin{table}[t]
\centering
\small
\setlength{\tabcolsep}{2.2mm}
\begin{tabular}{ccccccccc}
\toprule
\multirow{2}{*}{Data} & \multirow{2}{*}{Model Size} & \multicolumn{3}{c}{R Precision} & \multirow{2}{*}{FID$\downarrow$} & \multirow{2}{*}{MM Dist$\downarrow$} & \multirow{2}{*}{Diversity$\uparrow$} & \multirow{2}{*}{MModality$\uparrow$} \\
                      &                             & Top1$\uparrow$        & Top2$\uparrow$       & Top3$\uparrow$       &                      &                          &                            &                      \\ \midrule
20\%                  & \multirow{4}{*}{9 layer}    & 0.3379    & 0.5276   & 0.6381   & 7.7581               & 20.1603                  & 37.54                      & 38.17                \\
50\%                  &                             & 0.3664    & 0.5649   & 0.6889   & 7.0489               & 17.3070                   & 37.29                      & 36.77                \\
80\%                  &                             & 0.3697    & 0.5736   & 0.6902   & 6.5420                & 17.0101                  & 37.40                       & 38.01                \\
100\%                 &                             & 0.3761    & 0.5775   & 0.6932   & 3.9038               & 16.8211                  & 38.34                      & 38.63                \\ \midrule
20\%                  & \multirow{4}{*}{5 layer}    & 0.2965    & 0.4881   & 0.6127   & 22.4394              & 20.6333                  & 37.39                      & 38.20                 \\
50\%                  &                             & 0.3234    & 0.5179   & 0.6422   & 12.4010               & 18.8811                  & 37.54                      & 36.06                \\
80\%                  &                             & 0.3277    & 0.5226   & 0.6466   & 8.9618               & 19.1121                  & 37.67                      & 36.99                \\
100\%                 &                             & 0.3375    & 0.5384   & 0.6593   & 10.5134              & 18.4968                  & 37.71                      & 36.41                \\ \bottomrule
\end{tabular}
\captionsetup{skip=2pt}
\caption{Scaling-up performance analysis for Tender model on the (D1) subset across all performance metrics. Experimental findings underscore the robust scalability of our transformation-based Render model, particularly in its capacity to seamlessly align textual semantics with the synthesis of diverse whole-body human movements. }
\label{scaling-up-diffusion-more}
\end{table}

\begin{table}[t]
\small
\centering
\setlength{\tabcolsep}{1.9mm}
\begin{tabular}{lccccccc}
\toprule
\multirow{2}{*}{Method} & \multicolumn{3}{c}{R Precision} & \multirow{2}{*}{FID$\downarrow$} & \multirow{2}{*}{MM Dist$\downarrow$} & \multirow{2}{*}{Diversity$\uparrow$} & \multirow{2}{*}{MModality$\uparrow$} \\ 
                                       & Top1$\uparrow$        & Top2$\uparrow$       & Top3$\uparrow$       &                      &                            &                            \\ \midrule
Ours-9-1-256                                  & 0.2704    & 0.4381   & 0.5497   & 12.9417   &  22.7567         & 37.64                      & 37.51                      \\
Ours-9-2-256                                  & 0.2750    & 0.4318   & 0.5328   & 18.1709    &  24.1560         & 36.99                      & 38.09                      \\
Ours-9-4-256                                  & 0.3272    & 0.5053   & 0.6213   & 9.2707     &  20.9216         & 36.69                      & 38.81                      \\
Ours-9-8-256                                  & 0.3379    & 0.5267   & 0.6381   & 6.6322     &   20.1603   & 37.51                      & 34.79                      \\
Ours-9-16-128                                 & 0.2946    & 0.4693   & 0.5748   & 15.6112     &  22.2540         &  37.33                     & 38.14                      \\
Ours-9-16-64                                   & 0.2287    & 0.3594   & 0.4578   & 22.8346    &  27.0285         & 36.43                      & 37.37                      \\ \midrule
Ours-5-8-256                                 &  0.3351   &  0.5163  &  0.6265  &  8.5869    &   20.1136        &  38.70                     &  38.35                     \\
Ours-9-8-256                                 & 0.3379    & 0.5267   & 0.6381   & 6.6322     &   19.0618   & 38.61                      & 38.30                      \\
Ours-13-8-256                                 &  0.3227    & 0.5071   &  0.6184  &  9.1692   &  20.6883    &  36.93                     &   37.84                    \\  \midrule
Ours-9-8-256, w/ skip                                 & 0.3379    & 0.5267   & 0.6381   & 6.6322     &   20.1603   & 37.51                      & 34.79                      \\
Ours-9-8-256, w/o skip                                  & 0.3224     & 0.5196   &  0.6361  & 11.0998    &  20.4429 &  37.43                     &  38.19                     \\ \bottomrule
\end{tabular}
\captionsetup{skip=2pt}
\caption{Evaluation of our proposed PA-VAE module on the (\textbf{D3}) subset. The experimental setting as ``Ours-$l$-$n$-$d$'' specifies the structure of our part-aware motion encoder, where $l$ represents the number of transformer layers, $n$ and $d$ denotes the number and dimension of each latent code $x_z$. ``w/ skip'' and ``w/o skip'' denote the transformer layers of PA-VAE moudle with / without long skip connection operator.}
\label{Latent Dimension}
\end{table}

\begin{table}[t]
\centering
\small
\setlength{\tabcolsep}{2.4mm}
\begin{tabular}{lccccccc}
\toprule
\multirow{2}{*}{Text Condition} & \multicolumn{3}{c}{R Precision} & \multirow{2}{*}{FID$\downarrow$} & \multirow{2}{*}{MM Dist$\downarrow$} & \multirow{2}{*}{Diversity$\uparrow$} & \multirow{2}{*}{MModality$\uparrow$} \\
                                & Top1$\uparrow$        & Top2$\uparrow$       & Top3$\uparrow$       &                      &                          &                            &                      \\ \midrule
Concatenate                     & 0.3379    & 0.5276   & 0.6381   & 7.7581              & 20.1603                  & 37.54                      & 38.17                \\           
Add                             & 0.3070    & 0.4888   & 0.6101   & 11.5309              & 20.5131                  & 37.52                      & 37.57                \\
Text transformers               & 0.2662    & 0.4326   & 0.5444   & 12.1292              & 22.7878                  & 37.59                      & 37.35                \\
Cross Attention                 & 0.2900    & 0.4612   & 0.5719   & 19.6397              & 21.6411                  & 37.53                      & 36.27                \\ \bottomrule
\end{tabular}
\captionsetup{skip=2pt}
\caption{Comparative performance of different text conditioning methods on 2D whole-body motion generation. The \textit{concatenate} manner showing superior overall performance across all four methods.}
\label{text condition}
\end{table}

\begin{table}[t]
\centering
\setlength{\tabcolsep}{2.5mm}
\small
\begin{tabular}{cccccccc}
\toprule
                                 & \multicolumn{3}{c}{R Precision} &                       &                           &                             &                             \\
\multirow{-2}{*}{Designs of VAE} & Top1$\uparrow$        & Top2$\uparrow$       & Top3$\uparrow$       & \multirow{-2}{*}{FID$\downarrow$} & \multirow{-2}{*}{MM Dist$\downarrow$} & \multirow{-2}{*}{Diversity$\uparrow$} & \multirow{-2}{*}{MModality$\uparrow$} \\ \midrule
VAE                        & 0.3120    & 0.5079   & 0.6317   & 6.2446                & 16.1876                   & 37.78                       & 37.23                       \\
PA-VAE-2-4                       & \textbf{0.3250}    & \textbf{0.5210}   & \textbf{0.6509}   & \textbf{4.4032}                & \textbf{14.7318}                   & \textbf{37.86}                       & \textbf{38.49}                       \\
PA-VAE-2-3                       & 0.3215    &  0.5088        &  0.6255        &   7.0700                    & 15.9443                          & 37.64                            &  37.83                           \\
PA-VAE-2-2                       & 0.2636    & 0.4384   & 0.5568   & 6.3866                & 18.7368                   & 37.81                       & 37.05                       \\
PA-VAE-1-4                       & 0.2821    & 0.4547   & 0.5859   & 6.4368                & 18.2543                   & 36.99                       & 37.60                       \\
PA-VAE-1-2                       & 0.2524    & 0.4335   & 0.5569   & 20.0147               & 18.7878                   & 37.70                       & 37.11                       \\ \bottomrule
\end{tabular}
\captionsetup{skip=2pt}
\caption{Evaluation of various whole-body part-aware VAE configurations on the (\textbf{D2}) subset across all performance metrics. ``PA-VAE-2-4'' exhibits superior performance.}
\label{Part-aware VAE-more}
\end{table}

\begin{table}[t]
\small
\setlength{\tabcolsep}{2.85mm}
\centering
\begin{tabular}{cccccccc}
\toprule
                         & \multicolumn{3}{c}{R Precision} &                       &                           &                             &                       \\
\multirow{-2}{*}{Method} & Top1$\uparrow$        & Top2$\uparrow$       & Top3$\uparrow$       & \multirow{-2}{*}{FID$\downarrow$} & \multirow{-2}{*}{MM Dist$\downarrow$} & \multirow{-2}{*}{Diversity$\uparrow$} & \multirow{-2}{*}{MModality$\uparrow$} \\ \midrule
Only high                & 0.3761    & 0.5664   & 0.6774   & 3.9038                & 17.7588                   & 38.34                       & 37.73                 \\
Only high                & \cellcolor[HTML]{FFF2CC}{0.3892}    & \cellcolor[HTML]{FFF2CC}{0.5872}   & \cellcolor[HTML]{FFF2CC}{0.6950}   & \cellcolor[HTML]{FFF2CC}{2.4440}                & \cellcolor[HTML]{FFF2CC}{17.0250}                   & \cellcolor[HTML]{FFF2CC}{41.38}                       & \cellcolor[HTML]{FFF2CC}{41.28}                 \\ \midrule
Finetune                 & 0.3505    & 0.5431   & 0.6442   & 8.0525                & 19.6852                   & 41.56                       & 37.46                 \\
Argument-1             & 0.3658    & 0.5584   &  0.6674  &  7.9846    &   19.2769    &      40.99       &   37.86               \\
Argument-2             & 0.3728    & 0.5783   & 0.6995   & 6.5995                & 18.8519                   & 41.24                       & 38.16                       \\
Argument-4             & 0.3897    & 0.5811   & 0.6901   & 4.8562                & 16.2715                   & 41.52                       & 41.68                 \\
Argument-4             & \cellcolor[HTML]{FFF2CC}{0.3838}    & \cellcolor[HTML]{FFF2CC}{0.5907}   & \cellcolor[HTML]{FFF2CC}{0.7077}   & \cellcolor[HTML]{FFF2CC}{3.7428}                & \cellcolor[HTML]{FFF2CC}{16.1444}                   & \cellcolor[HTML]{FFF2CC}{42.08}                       & \cellcolor[HTML]{FFF2CC}{38.54}                 \\ \bottomrule
\end{tabular}
\captionsetup{skip=2pt}
\caption{Impact of low-quality motions on the (\textbf{D1}) subset performance. Rows highlighted represent the results from the MoLIP model, which incorporates the confidence of keypoints during training for enhanced precision.}
\label{low quality motion data-more}
\end{table}

\begin{table}[t]
\centering
\small
\setlength{\tabcolsep}{1.6mm}
\begin{tabular}{c|cc|ccccccc}
\toprule
\multirow{2}{*}{Velocity} & \multirow{2}{*}{Weighted} & \multirow{2}{*}{CAG} & \multicolumn{3}{c}{R-Precision} & \multirow{2}{*}{FID$\downarrow$} & \multirow{2}{*}{MM Dist$\downarrow$} & \multirow{2}{*}{Diversity$\uparrow$} & \multirow{2}{*}{MModality$\uparrow$} \\
                          &                           &                              & Top1$\uparrow$      & Top2$\uparrow$     & Top3$\uparrow$     &                      &                          &                            &                            \\ \midrule
                          &                           &                              & 0.3250    & 0.5210   & 0.6509   & 4.4032               & 15.5090                  & 37.86                      & 38.49                      \\
\checkmark                         &                           &                              & 0.3453    & 0.5536   & 0.6734   & 6.4038               & 14.4277                  & 37.16                      & 37.70                      \\
                          & \checkmark                         &                              & 0.3131    & 0.5196   & 0.6498   & 8.5602               &   15.7625                       & 38.03                      & 37.18                      \\
                          &                           & \checkmark                            & 0.3285    & 0.5253   & 0.6519   & 3.5675               & 15.2033                  & 36.13                      & 37.06                      \\
\checkmark                         & \checkmark                         &                              & 0.3244    & 0.5249   & 0.6475   & 10.1125              & 14.7794                  & 37.40                      & 37.69                      \\
\checkmark                         &                           & \checkmark                            & 0.3531    & 0.5568   & 0.6843   & 3.6970               & 14.0919                  & 40.17                      & 38.81                      \\
\checkmark                         & \checkmark                         & \checkmark                            &   0.3476 &      0.5418     &    0.6635     &    5.8864      &    14.6718     &      38.59        &    35.98                        \\ \bottomrule
\end{tabular}
\captionsetup{skip=2pt}
\caption{Ablations of the approaches to utilizing the whole-body confidence score on the (\textbf{D2}) subset across all performance metrics. ``Velocity'' denotes the Velocity loss used in training the PA-VAE module.}
\label{confidence-more}
\end{table}

\begin{table}[t]
\small
\setlength{\tabcolsep}{1.9mm}
\centering
\begin{tabular}{lccccccc}
\toprule
\multirow{2}{*}{Multi-domain Data} & \multicolumn{3}{c}{R Precision} & \multirow{2}{*}{FID$\downarrow$} & \multirow{2}{*}{MM Dist$\downarrow$} & \multirow{2}{*}{Diversity$\uparrow$} & \multirow{2}{*}{MModality$\uparrow$} \\ 
                                       & Top1$\uparrow$        & Top2$\uparrow$       & Top3$\uparrow$       &                      &                            &                            \\ \midrule
MLD (Only whole-body)
                                  & 0.3120    & 0.5079   & 0.6317   & 6.2446    &  17.4381         & 37.78                      & 37.23                      \\
Ours (Only whole-body)                                 & 0.3250    & 0.5210   & 0.6509   & 4.4032    &  15.5090      & 37.86 
                      & 38.49                      \\ \midrule
MLD  (w/ face)
                                  &  0.3299   &  0.5228  & 0.6412   &  7.2927   &  15.7578         &   41.20                    &   39.92                    \\
Ours (w/ face)                                 & 0.3443    & 0.5432   & 0.6731   & 4.1074     &   15.1652        & 38.81                      & 38.23                      \\ \midrule
MLD (w/ IDEA400)                                  &  0.3279   &  0.5278  & 0.6493   &  7.9774    &   15.2824        &  38.28     &  37.92                      \\
Ours (w/ IDEA400)  
                                  & 0.3388    & 0.5338   & 0.6541   & 3.6376     & 15.8301          & 39.17                      & 40.15                      \\
 \bottomrule
\end{tabular}
\captionsetup{skip=2pt}
\caption{Impact of multi-source data integration on 2D whole-body motion generation performance across all
performance metrics. Our Tender method showcases remarkable improvements upon the introduction of multi-source data.}
\label{Multi-Source Data Integration-more}
\end{table}

\textbf{UCF101}~\cite{soomro2012ucf101}, a widely employed video dataset, is meticulously designed for action recognition tasks. It comprises an extensive collection of 13,320 video clips, spanning 101 diverse action categories ranging from basketball shooting to guitar playing and makeup application. The dataset is characterized by its complex diversity, arising from variable backgrounds and visual noise induced by camera movements, occlusions, and background clutter. Homepage: \url{https://www.crcv.ucf.edu/data/UCF101.php}.

\textbf{Kinetics-400 (K400)}~\cite{kay2017kinetics} dataset is a substantial, high-quality collection of YouTube video URLs focusing on human-centric actions. K400 dataset encompasses 400 different categories of human actions, with at least 400 video clips for each category, each of which is approximately 10 seconds in length. The actions are centered around human activities and cover a broad range of categories, \textit{e.g.}, physical activities, skill-based actions, sports and fitness, creative and artistic actions. Homepage: \url{https://deepmind.com/research/open-source/kinetics}.

\textbf{Kinetics-700 (K700)}~\cite{kay2017kinetics} dataset is built on the foundation of K400, which expands the horizon of action recognition research with an impressive collection of around 650,000 video clips distributed across 700 diverse categories. Like K400, K700 are also sourced from YouTube and are trimmed to concise, action-rich sequences. K700 not only encompasses all categories from its predecessor but also introduces additional actions providing finer granularity and covering less common activities. Homepage: \url{https://deepmind.com/research/open-source/kinetics}.

\textbf{UBody}~\cite{lin2023one} dataset engineered for granular action recognition, documents activities involving the upper torso, capturing gestures, facial dynamics, and other movements above the waist. The variety of the video content, filmed in settings from domestic environments to offices and outdoor spaces. Homepage: \url{https://osx-ubody.github.io}.

\textbf{InternVid}~\cite{wang2023internvid} is a recently web-scale video-language multi-modal dataset for developing powerful and transferable video-text representations that are vital for multi-modal understanding. InternVid has 7 million videos, corresponding to 234 million clips and highly-correlated textual descriptions of total 4.1B words. These videos cover 16 scenarios and are around 6,000 motion descriptions. Homepage: \url{https://github.com/OpenGVLab/InternVideo/tree/main/Data/InternVid}.

\textbf{HAA500}~\cite{chung2021haa500} dataset stands as a meticulously curated repository for action recognition, distinguished by its human-centric focus on atomic actions. It features 500 classes with over 591,000 labeled frames. Each class in HAA500 focuses on fine-grained atomic actions, which means that only consistent actions fall under the same label, such as ``Baseball Pitching'' versus ``Free Throw in Basketball''. HAA500 contains a wide variety of atomic actions, ranging from athletic atomic action (Figure Skating - Ina Bauer) to daily atomic action (Eating a Burger). Homepage: \url{https://www.cse.ust.hk/haa}.

\textbf{Something-Something V2~(Sth-v2)}~\cite{goyal2017something} dataset features an extensive collection of 220,847 video clips distributed among 174 different action categories (e.g., Pushing, Tilting, Throwing, Tapping). These clips vividly document the interactions between humans and objects, including detailed directions and methods of these interactions, as illustrated by actions like ``Pushing something from left to right'' and ``Putting something on a surface''. Within our data processing pipeline, we integrate the textual labels from the Something-Something V2 dataset directly into ChatGPT~\cite{}, resulting in precise and comprehensive hand motion descriptions. Homepage: \url{https://developer.qualcomm.com/software/ai-datasets/something-something}.

\textbf{DFEW (Dynamic Facial Expressions in the Wild)}~\cite{jiang2020dfew} dataset is an extensive video repository curated for dynamic facial expression recognition. It includes 16,372 video clips, each with a duration of 1 to 6 seconds, covering a broad spectrum of facial expressions, including Anger, Disgust, Fear, Happiness, Sadness, Surprise, and Neutral. Leveraging ChatGPT, we transform facial expression labels into comprehensive, detailed descriptions of facial expressions. Homepage: \url{https://dfew-dataset.github.io}.

\textbf{CAER (Context-Aware Emotion Recognition)}~\cite{lee2019context} dataset includes a substantial number of video clips sourced from 79 different TV shows. The CAER dataset comprises a total of 13,201 video sequences, each approximately 90 frames in length, which is annotated with seven primary emotion categories: Happiness, Sadness, Surprise, Fear, Disgust, Anger, and Neutral. Homepage: \url{https://caer-dataset.github.io}.

\subsection{Usage and Distribution}
\label{Usage and Distribution}
\begin{itemize}[leftmargin=*]
\setlength{\leftmargin}{0pt}
    \item \textbf{Dataset Access Links.} The dataset is released at \url{https://holistic-motion2d.github.io}, which will be available during the review process and updated upon final publication.
    \item \textbf{Accessibility and Format.} The motion data is saved in \texttt{pkl} format and the textual description is saved in \texttt{txt} format, where an example is shown in the README.md file. An example code snippet is also provided showing how to read and process the data. The dataset will be hosted on the Huggingface platform with a commitment to maintain long-term availability.
    \item \textbf{Data License Confirmation and Author Responsibility.} All the \textbf{Holistic-Motion2D} is distributed under the CC-BY-NC-SA (Attribution-NonCommercial-ShareAlike) license to ensure its legitimate and widespread use. For the sub-datasets of \textbf{Holistic-Motion2D}, we would ask the user to read the original license of each original dataset, and we would only provide our annotated result to the user with the approvals from the original Institution. We assume all responsibilities for potential legal issues arising from the use of the dataset. We confirm that our Holistic-Motion2D does not contain any personally identifiable information or offensive content.
    \item \textbf{Code License}. The code for pre-processing and training our Tender model uses the MIT license. Please refer to the GitHub repository for license details.

\end{itemize}

\subsection{Visualization of Motion and Caption}

In this section, we present more visual samples of the original data from the test set of the \textbf{\DatasetName}. Our motion visual results of the \textbf{\DatasetName} are shown in Figure ~\ref{fig:visual_results}. By comparing the 2D poses processed through the motion annotation pipeline with the descriptions processed through the text annotation pipeline, we found that our annotation method captures accurate motion information from both pose and text perspectives, even in different visual scenes.
Furthermore, we show some cases about the pipeline of annotating the captions in Figure \ref{fig:caption_visualization}.

\section{Additional Experiment}
\subsection{Experimental Setting}
\label{sec:additional_experimental_setting}
In our work, the training of the proposed Tender unfolds in two distinct stages to optimize performance and accuracy in generating 2D whole-body motions. Firstly, the Part-Aware Variational Auto-Encoder (PA-VAE) is trained to effectively capture the whole-body spatio-temporal motion representations. Secondly, we train the transformer-based diffusion model within the latent space. All training stages integrates the confidence scores of keypoints via the Confidence-Aware Generation (CAG) mechanism, gaining improved perceptual abilities to prioritize the reliability of 2D keypoints.

The motion encoders $\mathcal{E}$ and decoders $\mathcal{D}$ of our proposed PA-VAE model are all composed of 9 skip-connected transformer layers along with 4 heads. In terms of the transformer-based denoiser, the text embedding $\phi_\theta(c)\in\mathcal{R}^{1\times 256}$ and the motion latent $x_{z}\in\mathcal{R}^{8\times 256}$ are concatenated for diffusion learning and inference. For the text conditioning, we utilize a pre-trained CLIP \cite{radford2021learning} ViT-L/14@336px~\cite{dosovitskiy2020image} as the default text encoder. Additional text injection methods are presented in Table~\ref{text condition} to demonstrate their impact through ablation experiments. Our mini-batch size on each GPU is set to 32 during the PA-VAE training stage and 256 during the diffusion training stage. All our models are trained with the AdamW~\cite{loshchilov2017decoupled} optimizer using a fixed learning rate of 2e-4 without weight decay. The code of our proposed Tender is implemented on the PyTorch~\cite{paszke2017automatic} platform equipped with 4 NVIDIA 80G A100 GPUs on internal clusters for the PA-VAE and diffusion model training. We train our model for 1,500 epochs during the reconstruction stage of VAE and 2,000 epochs during denoising stage of diffusion. Similar to the setting of MLD~\cite{chen2023executing}, the number of diffusion steps is 1,000 during training while 50 during interfering, and the variances $\beta_t$ are scaled linearly from $8.5\times10^{-4}$ to 0.012. Further, the scale $s$ in classifier-free diffusion guidance is set to 7.5.
The high quality motions selected by us satisfy two rules: 1) the length of motion longer than 64; 2) the number of keypoints with confidence higher than 0.3 larger than 70 percent of total keypoints.

During training for evaluation model, we use ViT-B/16 \cite{dosovitskiy2020image} pre-trained by MAE \cite{he2022masked} for motion encoder $\mathcal{E}_m$ to help convergence.
And we use the text encoder of CLIP \cite{radford2021learning} ViT-L/14@336px \cite{dosovitskiy2020image} as our text encoder backbone, which is fixed during training.
We append 2 transformer \cite{vaswani2017attention} layers and a LayerNorm \cite{ba2016layer} layer after this backbone to help the retrieval model learn text-motion aligned feature representations, which are trainable.
The batch size $N$ and motion length $L$ are set as 1024 and 200, respectively.
Following HumanML3D \cite{guo2022generating}, we use the following evaluation metrics in our benchmark: R-Precision (Top1/2/3), Frechet Inception Distance (FID), Multimodal Distance (MM-Dist), Diversity, and Multimodality (MModality).

\begin{figure}[t]
    \begin{center}
        \centerline{\includegraphics[width=\linewidth]{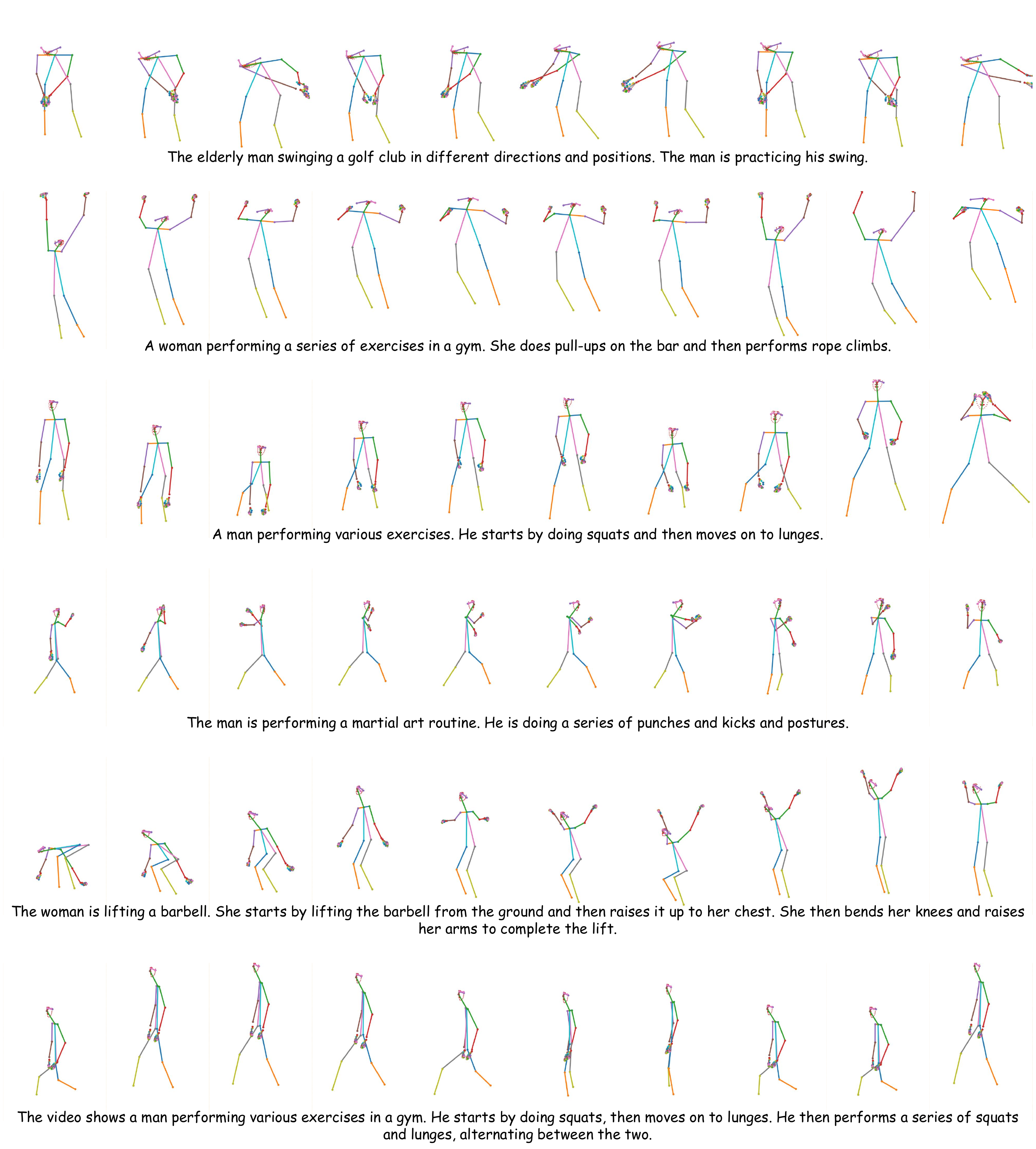}}
        \caption{Visualization results of the generated 2D whole-body motions using our proposed \textbf{\MethodName} model in multiple visual scenarios, with a corresponding text prompt given below.} %
        \label{fig:visual_results_model}
    \end{center}
\end{figure}

\begin{figure}[t]
    \begin{center}
        \centerline{\includegraphics[width=\linewidth,height=0.9\textwidth]{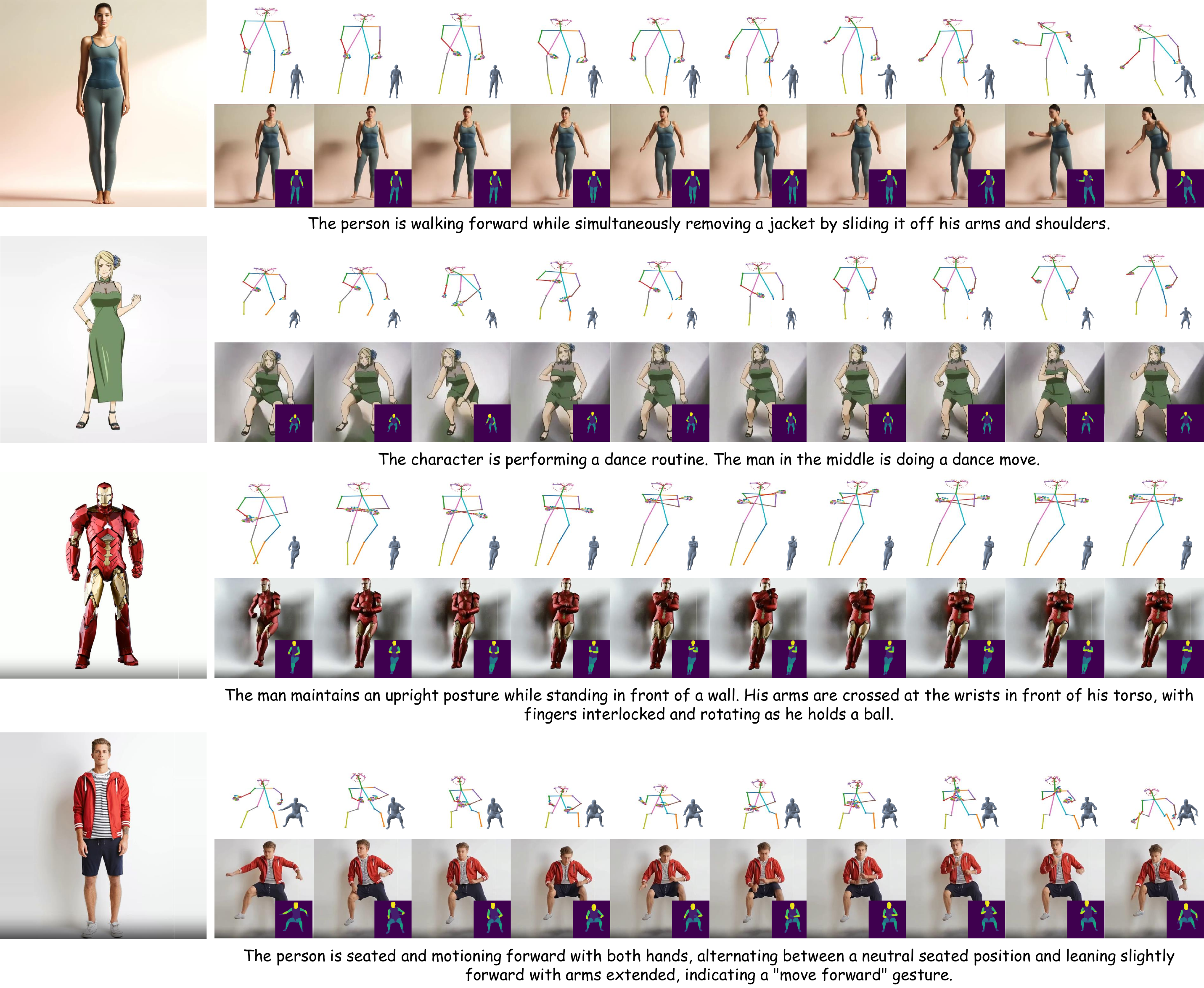}}
        \caption{Visualization results of pose-guided human video generation using our proposed \textbf{\MethodName} model in multiple visual scenarios, with a corresponding text prompt given below and a reference image given on the left.} %
        \label{fig:appendix_human_video}
    \end{center}
\end{figure}

\begin{figure}[t]
    \begin{center}
        \centerline{\includegraphics[width=\linewidth]{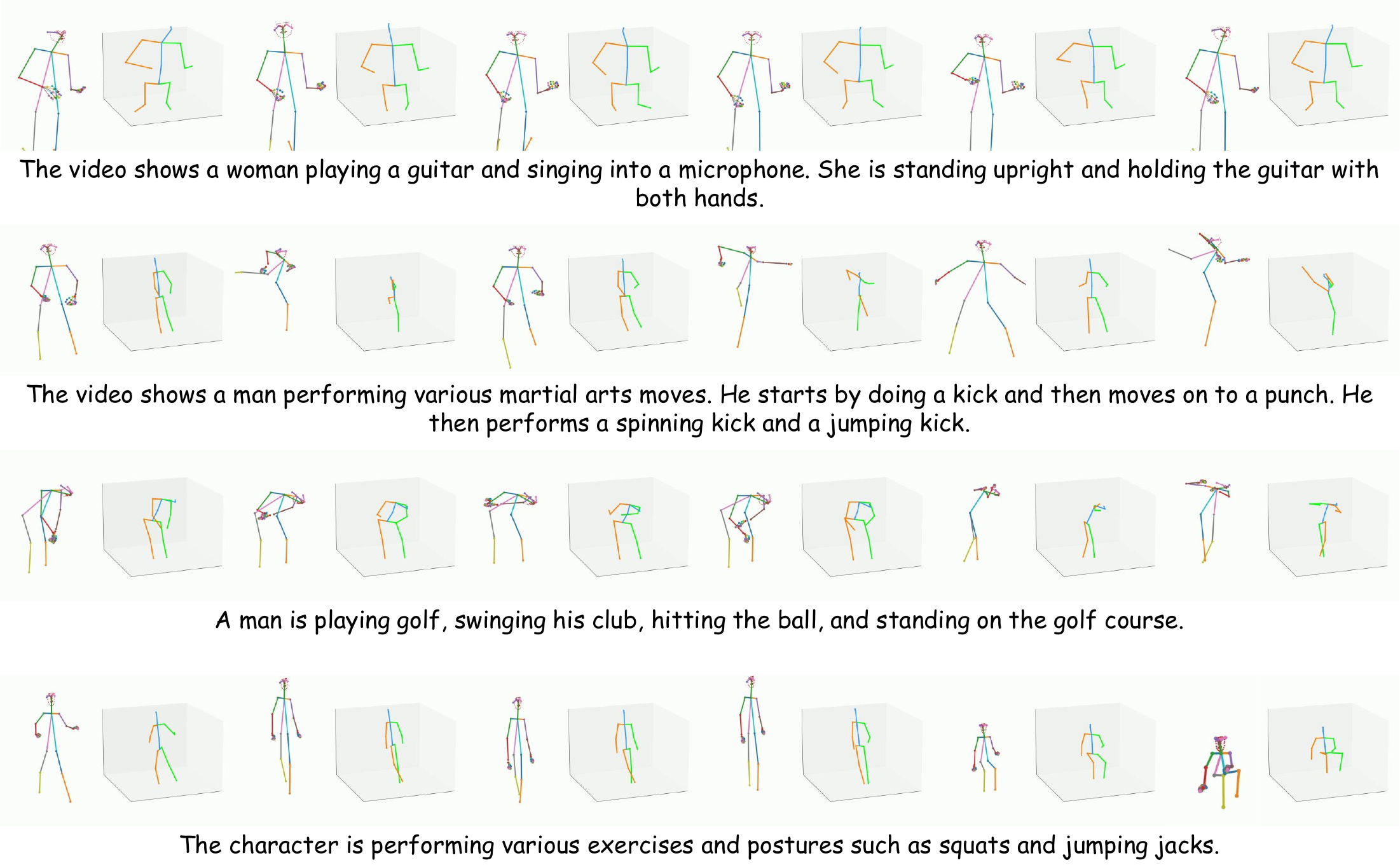}}
        \caption{Visualization results of 3D motion lifting using our proposed \textbf{\MethodName} model in multiple visual scenarios, with a corresponding text prompt given below.} %
        \label{fig:3d_motion_lifting}
    \end{center}
\end{figure}


\subsection{More Quantitative Results}

\paragraph{More Quantitative Results of the Scaling Law Experiments.} In this part, we extend Table~\ref{Scaling law} and Table~\ref{tab:scaling_law_t2m_model} and provide more quantitative results of our MoLIP and Tender's scaling-up capability. The Table~\ref{Scaling law-more} provides an in-depth scaling-up performance analysis of the MoLIP model on the (D1) subset, quantifying its ability to align text with generated motion across increasing dataset sizes. The reduction in the standard deviations across all metrics as data increases demonstrates the robustness of the MoLIP model, confirming its stability and reliability in producing consistent results (e.g. R Precision, Diversity, and FID) over multiple evaluations. 

\paragraph{Effectiveness of Whole-body Motion Latent Representations.} In Table~\ref{Latent Dimension}, we evaluate some key parameters choice for the PA-VAE module. The latent vector $x_{z}\in\mathcal{R}^{n\times d}$ stands as the most crucial variable in Tender-2D, where $n$ and $d$ denote the number and dimensions of $x_{z}$. We formulate the experimental setting as ``Ours-$l$-$n$-$d$'', where $l$ is the number of temporal transformer layers. Further, we find skip connections mechanism prove beneficial in facilitating the learning of latent representations for 2D whole-body motion. As shown in Table~\ref{Latent Dimension}, the Tender-2D with latent vector $x_{z}\in\mathcal{R}^{8\times 256}$, 9 temporal transformer layers, and skip connection demonstrates superior performance.

\textbf{Ablation Study of the Text Condition}. Among all tested method in Table~\ref{text condition}, concatenating the text features with  approach notably outperforms others in synthesizing 2D whole-body motions. Incorporating text transformers subsequent to the text embedding (\textit{Text transformers}) and \textit{cross attention} does not effectively translate textual information into controllable signals for 2D motion generation, manifesting suboptimal performance.

\paragraph{Other Holistic Experiments Results.} In response to space limitations in the main document, where only a subset of evaluation metrics from our ablation studies could be presented, this section provides a detailed compilation of all results across various evaluation metrics. We detail the Ablation Studies of the PA-VAE moudle in Table~\ref{Part-aware VAE-more}, Impact of low-quality motions in Table~\ref{low quality motion data-more}, Approaches to Utilizing Confidence in Table~\ref{confidence-more}, and Multi-source Datasets experiments in Table~\ref{Multi-Source Data Integration-more}. The manuscript offers detailed experimental analyses to substantiate the effectiveness of the proposed Tender model in generating high-quality, expressive, and diverse whole-body human motions.

\subsection{More Visualization Results} 
\paragraph{2D Whole-body Human Motions.}
Our qualitative evaluation, as depicted in Figure~\ref{fig:visual_results_model}, provides a compelling visual demonstration of the capabilities of the proposed baseline method \textbf{\MethodName}. The visualizations of 2D poses generated by our model vividly illustrate its remarkable adaptability across a diverse range of application scenarios. Notably, the generated 2D motion sequences are characterized by their spatial and temporal coherence, ensuring that movements appear smooth and natural over time. Furthermore, these sequences exhibit exceptional alignment with descriptive texts, showcasing ability to accurately interpret complex textual commands into motion dynamics.

\paragraph{Downstream Applications.}
We also show additional visualization results for pose-guided human video generation and 3D motion lifting in Figure \ref{fig:appendix_human_video} and Figure \ref{fig:3d_motion_lifting}, respectively.
It is learned that high-quality and consistent human videos and 3D human motions can be effectively generated with 2D human motions produced by our proposed \MethodName method. These results further demonstrate the strong generalization capability and robustness of our approach.

\section{Limitation and Broader Impact}

\subsection{Limitation}
\label{Limitation}
While our work on the Holistic-Motion2D dataset and the Tender-2D model represents significant advancements in the field of 2D whole-body motion generation, there are several limitations that must be acknowledged: 1) \textbf{Single-Person Motion Synthesis}. Our proposed model is designed to generate motions for an individual person only and cannot support scenarios involving multiple persons. This restriction significantly limits the ability to model complex interactions that are typical in real-world settings, \textit{e.g.}, social gatherings, celebratory events or group activities, where interactions are crucial. Given our principal focus is the development of a 2D single-person motion generation dataset and benchmark, we have identified this as an area for future investigation. 2) \textbf{Dependency on Annotation Quality}. The effectiveness of our Tender-2D model heavily relies on the quality of the annotations provided in the Holistic-Motion2D dataset. Inaccuracies in motion annotations or textual descriptions compromise the training process, producing generated motions that deviate from the intended actions.

\subsection{Broader Impact}

Our contribution through the development of the Holistic-Motion2D dataset substantially enhances the corpus of academic resources, furnishing researchers with access to over a million motion sequences with detailed annotations. Such a large-scale 2D whole-body motion generation dataset ushers in a new era for the synthesis of 2D whole-body motions, establishing a robust foundation that not only facilitates advanced research in traditional areas of motion analysis but also propels forward cutting-edge applications in AI-driven content creation, such as AR/VR and Human-Machine Interaction. To date, human motion datasets have not been associated with any direct negative social impacts. Our proposed Holistic-Motion2D  will strictly follow the license of previous datasets, and would not present any negative foreseeable societal consequence, either.

{\small
\bibliographystyle{unsrt}
\bibliography{neurips_data_2024}
}

\end{document}